%% file: iclr2027_conference.tex
\documentclass{article} 
\usepackage{iclr2027_conference,times}

\input{math_commands.tex}

\usepackage[utf8]{inputenc} 
\usepackage[T1]{fontenc}    
\usepackage{url}            
\usepackage{booktabs}       
\usepackage{subcaption}     
\usepackage{amsfonts}       
\usepackage{nicefrac}       
\usepackage{microtype}      
\usepackage{xcolor}         
\usepackage{mathtools}      
\usepackage{amsmath,amssymb,mathrsfs,bm,latexsym,graphicx}
\usepackage{multirow}
\usepackage{paralist}
\usepackage{setspace,color,verbatim}
\usepackage{algorithm}
\usepackage{algpseudocode}
\usepackage{graphicx}
\usepackage{amsthm}

\usepackage{dsfont}

\usepackage{tabularx}
\usepackage{array}
\usepackage{hyperref} 
\usepackage{cleveref} 
\usepackage{bbm} 
\usepackage{makecell} 
\usepackage{pifont}

\newcommand{\method}{\textsc{SelfMorph}}
\newcommand{\basecomp}{\textsc{Base}}
\newcommand{\rep}{\textsc{Repeat}}
\newcommand{\branch}{\textsc{Branch}}

\title{Compute Has Shape: Self-Morphing the \\Internal Computation of Frozen LLMs}

\author{Junxi Wu$^{1,2}$~,\quad Yiwen Guo$^{3}$\thanks{Corresponding authors.}~ , \quad Shu-Tao Xia$^{2}$\footnotemark[2]{}~\\
$^{1}$ School of Statistics and Data Science, Nankai University \\
$^{2}$ Tsinghua Shenzhen International Graduate School, Tsinghua University \\
$^{3}$ Independent Researcher}
\iclrfinalcopy 

\begin{document}

\maketitle
\lhead{Preprint}

\maketitle

\input{Sections/abstract}

\input{Sections/introduction}

\input{Sections/method}
\input{Sections/experiment}

\input{Sections/conclusion}


\bibliography{iclr2027_conference}
\bibliographystyle{iclr2027_conference}

\newpage

\appendix
\crefalias{section}{appendix}
\crefalias{subsection}{appendix}
\input{Sections/appendix}

\end{document}

%% file: math_commands.tex
\usepackage{amsmath,amsfonts,bm}

\def\eqref#1{equation~\ref{#1}}

\def\1{\bm{1}}

\DeclareMathAlphabet{\mathsfit}{\encodingdefault}{\sfdefault}{m}{sl}
\SetMathAlphabet{\mathsfit}{bold}{\encodingdefault}{\sfdefault}{bx}{n}

\DeclareMathOperator*{\argmax}{arg\,max}

%% file: Sections/abstract.tex
\begin{abstract}

Test-time scaling improves large language models through additional computation, but how to adapt the topology of internal computation to individual inputs remains underexplored. Existing approaches largely fix recurrent or parallel architectures during training, or reconfigure frozen models along a single structural axis. We introduce \method{}, which makes computational shape an input-dependent inference-time decision. It unifies the original forward path (\basecomp{}), serial reuse of a layer span (\rep{}), and local branching and merging of perturbed hidden states (\branch{}) in a unified space parameterized by type, location, and intensity. Since they offer complementary benefits rather than uniform dominance, we use a lightweight selector that reads the backbone’s depth-wise hidden state trajectory on the prompt to choose a morphology before response generation. We train the selector by distilling utility-based distribution from offline morphing that jointly account for task performance and computational cost. 
Evaluated on three LLMs spanning 4B–27B, \method{} achieves a 8.06\% mean relative improvement on five in-distribution benchmarks and 4.45\% on four unseen benchmarks. It achieves the strongest performance in 24 of 27 settings. These findings identify input-adaptive computational topology as a complementary axis of test-time scaling beyond generating more tokens. Our code is available at https://github.com/creator-xi/SelfMorph.
\end{abstract}

%% file: Sections/introduction.tex
\section{Introduction}
\label{sec:introduction}

Scaling test-time computation has become a important way to improve large language models. Chain-of-thought increases serial computation by generating longer reasoning traces \citep{wei2022chain}, while self-consistency increases parallel computation by sampling multiple traces and aggregates answers \citep{wang2022self}. These approaches organize additional computation through generated sequences. Computation can also be allocated within the network, raising a further question: \emph{can a frozen LLM adapt its internal computation to each input by choosing among serial reuse, local parallel processing, and its original execution path?}

Recent work provides two foundations for this question. Looped (recurrent-depth) models repeatedly apply shared transformations to deepen serial processing \citep{dehghani2019universal,geiping2026scaling,zhu2025scaling}, while parallel models maintain multiple latent streams and aggregate their outputs \citep{chen2026parallel,you2026parallel}. These studies establish mechanisms for organizing computation within models designed or trained to support them.
Another line of work shows that the pretrained layers can be skipped or repeated at inference time \citep{li2026skip,chen2026training}, demonstrating that frozen weights can support more than one execution path. 
However, these methods operate within a single axis, and do not place different topologies under a shared cost model. This motivates a broader view: \textbf{test-time computation has shape, not only scale}. We investigate whether these computational shapes offer complementary benefits and whether the backbone's own hidden state trajectory can guide their selection.

\begin{figure}[t]
    \centering
    \vspace{-2em}
    \includegraphics[width=\linewidth]{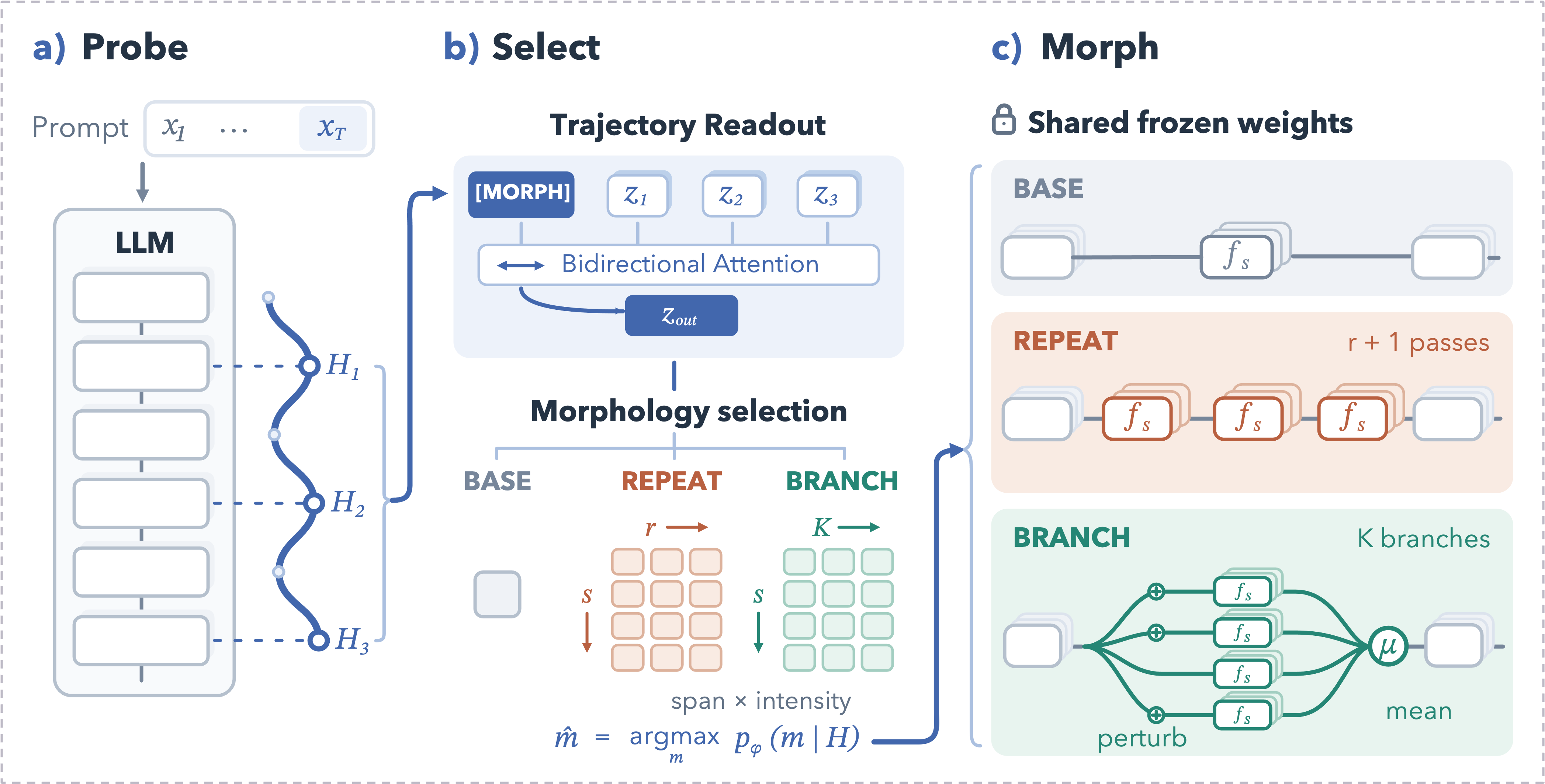}
    \caption{
        Overview of \textsc{SelfMorph}.
        (a) Probe: During prompt prefill along the original forward
        path, we collect a depth-wise trajectory.
        (b) Select: A lightweight selector encodes the trajectory
        and aggregates them through bidirectional
        attention with a learned token. Before response
        generation, it selects the highest-probability action.
        (c) Morph: Three morphology types.
    }
    \label{fig:framework}
    \vspace{-1em}
\end{figure}

We formalize computational shape through a parameterized morphology space over a frozen backbone. \basecomp{} preserves the original forward path; \rep{} applies a selected layer span repeatedly in series; \branch{} forks perturbed hidden states at the start of the span and merges at the end. The morphology type describes the topology of the computation, while the intervention span and intensity further specify where and how much of the computation is applied. 
Evaluating these interventions on the same problems reveals complementary effects: \rep{} and \branch{} each solve problems that the other does not, and neither uniformly subsumes the other across tasks and backbones. This heterogeneity motivates selecting a morphology for each input rather than applying a single fixed intervention.

As the model encodes the prompt, its hidden state traces a depth-wise trajectory across layers. We introduce \method{}, which reads the trajectory of the LLM itself to decide the shape it ought to adopt. 
Specifically, \method{} predicts the parameterized morphology, including type, span, and intensity before the first response token is generated. 
To train the selector, we evaluate candidate actions offline and assign each a utility combining task score and analytic cost. We distill the resulting utility-based teacher distribution into the selector.
At inference time, the selected morphology reshapes the model's internal computation throughout generation without evaluating candidate responses. To reduce execution overhead, we reuse prompt computation shared by probing, and share identical cache storage across branch replicas where supported.

Our contributions are as follows.
\begin{itemize}
    \item We formulate the internal computation of a frozen LLM as a structured choice over \basecomp{}, \rep{}, and \branch{}, jointly parameterized by morphology type, span, and intensity. This formulation places serial reuse and local branching within a shared action space over pretrained weights.

    \item We introduce \method{}, a lightweight attention-based selector that reads the frozen LLM's hidden state trajectory on the prompt and predicts which morphology to apply before generation. We train the selector by distilling utility-based distribution from offline morphing that jointly account for task performance and computational cost.

    \item We evaluate \method{} on three frozen LLMs across nine benchmarks. 
    Relative to the unmodified backbones, it achieves a mean relative score improvement of 8.06\% on five in-distribution benchmarks and 4.45\% across on four unseen benchmarks. It also outperforms a per-benchmark selected fixed action in all 15 settings, supporting the value of input-level selection.
    Extensive ablations also confirm the contributions of morphology diversity, hidden trajectory and attention-based architecture.

\end{itemize}

%% file: Sections/method.tex
\section{Parameterized Morphology Space}
\label{sec:morphology-space}

LLMs can be morphed by choosing \emph{what} additional
computation to perform (the morphology type), \emph{where} in the layer
stack to apply it (the span), and \emph{how much} to add (the
intensity). Together they define a parameterized morphology space over a frozen backbone.

\subsection{Morphology Types}
\label{sec:method-space}

Let $\mathcal{M}$ denote a frozen LLM with $N$ layers. A half-open
span $s{=}(i,j)$ covers layers $i, i{+}1, \dots, j{-}1$ with
$0 \le i < j \le N$. We denote $h_i$ for the hidden state entering layer $i$ and $f_s = f_{j-1} \circ\cdots\circ f_i$ for the composite transformation over the span. We define three morphology types below, each using the same frozen weights but organizing the computation through $s$ differently.

\paragraph{\basecomp{}.}
The span is traversed once: $h_j^{\basecomp{}} = f_s(h_i)$. No extra
computation is added.

\paragraph{\rep{}.}

The span is traversed $r{+}1$ times in series, where the integer $r\geq 1$
specifies the number of additional passes. Each pass taking the
output of the previous one as input:
\[
h_j^{\rep{}} = f_s^{\,r+1}(h_i), \qquad
f_s^{\,n} := \underbrace{f_s\circ\cdots\circ f_s}_{n\text{ times}}.
\]
Successive passes build on earlier updates, providing additional
opportunities to refine features before passing the state onward.
Each layer retains its built‑in normalization throughout the repeated execution.
In this sense, \rep{} lets the model ``think longer'' within a selected layer span, increasing effective depth without introducing new parameters.

\paragraph{\branch{}.}

The input state is replicated into $K\geq 2$ branches, where $K$ is
an integer. At the span entrance, each branches receives a perturbation along an independently sampled Gaussian direction:
$$
\tilde h_i^{(k)} = h_i + \sigma_k \lVert h_i \rVert_2\, \hat\epsilon^{(k)},
\qquad \sigma_k = \frac{k}{K-1}\sigma_{\max}, \quad k=0,\dots,K{-}1,
$$
where $\hat\epsilon^{(k)} = \epsilon^{(k)}/{\lVert\epsilon^{(k)}\rVert_2}$, $\epsilon^{(k)} \overset{\mathrm{i.i.d.}}{\sim}
\mathcal{N}(\boldsymbol{0},\boldsymbol{I}_{d_h})$ and $d_h$ is the dimension of hidden state. 
The hyperparameter $\sigma_{\max}$ controls the maximum
relative perturbation magnitude. The schedule preserves one unperturbed
input ($\sigma_0=0$) and scales the remaining perturbations relative to
the input state's norm. Each branches traverses the span independently,
and the resulting states are merged by arithmetic averaging:
$$
h_j = \frac{1}{K}\sum_{k=0}^{K-1} h_j^{(k)}, \qquad
h_j^{(k)} = f_s\bigl(\tilde h_i^{(k)}\bigr).
$$
The merged state is passed to the remaining layers.
\branch{} can be seen as exploration in hidden space: controlled
perturbations open alternative computational trajectories through the
span, creating opportunities to move beyond the original trajectory
before the branch outputs are reconciled.
The unperturbed branch provides a reference within the ensemble,
anchoring this exploration by preserving a direct contribution from
the unperturbed input.

Together, these morphologies expose two distinct ways to use
additional internal computation: \rep{} deepens the processing of an
evolving representation, while \branch{} broadens the computation
through perturbation and aggregation.

For each non-\basecomp{} morphology, an \emph{action} is a tuple $m = (\tau, s, \theta)$, where $\tau \in \{\rep{}, \branch{}\}$ specifies
the morphology type, $s$ specifies the location span, and $\theta$ is $r$ for \rep{} or $K$ for \branch{}. 
We represent \basecomp{} as a single action with no span or intensity choice.

\subsection{Constructing the Action Space}
\label{sec:method-family}

The morphology space contains $O(N^2)$ candidate spans for each
intervention type, each of which can be paired with multiple repetition
or branch counts. 
Moreover, intervention benefits vary substantially across this space, as many span configurations yield negligible gains or degrade performance.
We therefore construct a compact action space through screening on two diagnostic datasets, numerical estimation and emotion-intensity prediction,
which are separate from the evaluation benchmarks. At a fixed intensity for each type, we evaluate every feasible span and measure its score gain over \basecomp{}.

The screening reveals distinct depth preferences across morphology types.
Favorable \rep{} spans concentrate at intermediate
depths, whereas favorable \branch{} spans tend to occur earlier.
This separation is more distinct on the larger backbones.
This asymmetry is consistent with the functional progression across
model depth described in prior work: shallow layers build contextual
representations, middle layers process reasoning and refine abstract thought,
and later layers specialize toward token prediction
\citep{tenney2019bert,lad2025remarkable}.
Repeating an intermediate span provides additional opportunities to
refine features formed by preceding layers. Early branching instead
introduces variation before subsequent processing, allowing different
computational trajectories to develop within the span before their
outputs are merged and passed onward.

To select these candidates, we standardize the mean gains across spans
separately for each datasets, type, and backbone.
We rank spans by the sum of their standardized
gains on the two diagnostic datasets. After evaluating the leading candidates on a larger sample of the same diagnostic tasks, we retain four spans
per intervention type. Appendix~\ref{app:scan} provides the complete
screening and validation procedure, depth profiles, and selected spans.

For a given backbone, let $\mathcal{S}_{\tau}$ denote the
spans for intervention type $\tau$, and $\Theta_{\tau}$ denote
its intensity grid. The candidate actions for each type form
$\mathcal{A}_{\tau}
=\{\tau\}\times\mathcal{S}_{\tau}\times\Theta_{\tau}$.
The full action space is
\begin{equation}
\label{eq:action-space}
\mathcal{A}
=
\{\basecomp{}\}
\cup \mathcal{A}_{\rep{}}
\cup \mathcal{A}_{\branch{}}.
\end{equation}
We use $\Theta_{\rep{}}=\{1,2,3\}$ and
$\Theta_{\branch{}}=\{2,4,8\}$, yielding 25 actions per backbone.
The resulting pool preserves variation in type, location, and intensity
while making exhaustive offline evaluation feasible.


\section{\method{}}
\label{sec:selfmorph}

With the morphology space defined, we now have a compact action space $\mathcal{A}$. The remaining question is: which action in $\mathcal{A}$ should be applied for a given state? Enumerating every action incurs an prohibitive computational cost. This section introduces a lightweight selector that makes this decision from the backbone's own hidden states.

\subsection{Utility-Based Distillation}
\label{sec:method-objective}

\paragraph{Action utility.}
For an input $x$ and an action $m\in\mathcal{A}$, we define the utility
as
\begin{equation}
\label{eq:utility}
U(x,m) = \mathrm{score}(x,m) - \lambda\big(C(m)-1\big),
\end{equation}
where $\mathrm{score}(x,m)\in[0,1]$ is the observed task score, $C(m) \ge 1$ is the analytic cost proxy, and $\lambda\geq 0$ controls the penalty for additional computation. The oracle action is
\begin{equation}
\label{eq:objective}
m^*(x) = \argmax_{m\in\mathcal{A}} U(x,m).
\end{equation}
Evaluating this choice requires generating and scoring candidate
responses, so it is available only offline.

\paragraph{Analytic cost.}

We assign each action a resource penalty based on its intervention width and intensity. For non-\basecomp{} actions,
\[
C(m) = 1 + \frac{j-i}{N}\cdot k_{\mathrm{eff}}(\tau,\theta),
\]
where $j{-}i$ is the span width, $N$ is the backbone depth, and
$k_{\mathrm{eff}}(\tau,\theta)$ is a coefficient determined by the
morphology type and intensity (for \rep{}, $k_{\mathrm{eff}} = r$; for
\branch{}, $k_{\mathrm{eff}}$ accounts for the effective parallel
overhead of $K$ branches).
We use an analytic proxy rather than wall-clock measurements
because runtime depends on hardware, generation length, and batching
strategy, all of which introduce noise unrelated to the morphology
itself.
Appendix~\ref{app:speed} reports the coefficients, their measurement basis, and resource measurements.

\paragraph{Teacher distribution.}
For each training input, we execute all actions in $\mathcal{A}$ and
record their utilities. We construct a teacher distribution
that balances utility with entropy:
\begin{equation}
\label{eq:teacher-objective}
q_T(\cdot\mid x)
=\operatorname*{arg\,max}_{q\in\Delta(\mathcal{A})}
\left\{\sum_{m\in\mathcal{A}}q(m)U(x,m)
+T\operatorname{Ent}(q)\right\},
\end{equation}
where $\Delta(\mathcal{A})$ is the probability simplex,
$\operatorname{Ent}(q)=-\sum_m q(m)\log q(m)$, and $T>0$.
The solution is the temperature-scaled softmax \citep{bogacz2017tutorial}:
\begin{equation}
\label{eq:teacher-distribution}
q_T(m\mid x)
=\frac{\exp\bigl(U(x,m)/T\bigr)}
{\sum_{m'\in\mathcal{A}}\exp\bigl(U(x,m')/T\bigr)}.
\end{equation}
Smaller $T$ concentrates the teacher on higher-utility actions, whereas
larger $T$ spreads probability more broadly. This construction gives
equal weight to exact ties and represents graded preferences among
actions, retaining alternatives that a single oracle label would discard.

\paragraph{Distillation objective.}
The selector learns to predict the teacher distribution using only
$H(x)$. We minimize the cross-entropy
\begin{equation}
\label{eq:distillation}
\mathcal{L}
=-\mathbb{E}_{x\sim\mathcal{D}_{\mathrm{train}}}
\left[\sum_{m\in\mathcal{A}}q_T(m\mid x)
\log p_{\phi}(m\mid H(x))\right].
\end{equation}
For the hierarchical prediction heads defined below, this loss
decomposes exactly into a type cross-entropy and conditional
span--intensity cross-entropies. Each conditional term is weighted by
the teacher's probability mass on its morphology type. Thus, the full
action distribution supplies supervision to both levels of prediction.

\subsection{Trajectory-Conditioned Selection}
\label{sec:method-selector}

\paragraph{Hidden trajectory.}
The selector must decide which morphology to apply before any morphology
has been executed. An intuitive approach makes decisions based on the input prompt. But there is a more fundamental signal: the model’s own intermediate computations \citep{azaria2023internal}. As LLMs process the input, the hidden state evolves across depth. 
The depth-wise trajectory exposes how the representation changes through
the network, providing features for predicting morphology.

\paragraph{Trajectory encoding.}
\label{sec:method-utility}

During prompt prefill, we collect
the last-token hidden state at $L$ preselected depths and stack these
vectors into $H\in\mathbb{R}^{L\times d_h}$. Each slice $H_l$ is
pre-normalized using corresponding RMS normalization from the backbone, then mapped through a learnable projection:

$$
z_l = W_{\mathrm{enc}}\,\mathrm{PreNorm}(H_l) + e_l,
\qquad l = 1,\dots,L,
$$
where $W_{\mathrm{enc}} \in \mathbb{R}^{d_z \times d_h}$ is shared
across all $L$ layers and $\{e_l\}$ is a learned embedding table that
provides the only positional information in the network.

\paragraph{Depth aggregation.}

We prepend a learned query token $[\mathrm{MORPH}]\in\mathbb{R}^{d_z}$
to the sequence of layer features. A single bidirectional Transformer
block allows this query to attend to all observed depths. Its output
at the query position, followed by LayerNorm, gives the shared decision
representation $z_{\mathrm{out}}\in\mathbb{R}^{d_z}$.

\paragraph{Hierarchical action prediction.}

A linear head produces the morphology type logits over $\{\basecomp, \rep, \branch\}$.
For each non-\basecomp{} type, a two-layer MLP maps $z_{\mathrm{out}}$
to $S \times P$ joint distribution over span and intensity, using softmax over $S{\cdot}P$ combinations. The outputs are:
\begin{equation}
\label{eq:action-distribution}
p_{\phi}(m\mid H)=
\begin{cases}
p_{\phi}(\basecomp{}\mid H), & m=\basecomp{},\\
p_{\phi}(\tau\mid H)\,
p_{\phi}(s,\theta\mid\tau,H), & m=(\tau,s,\theta).
\end{cases}
\end{equation}
The factorization shares a trajectory representation across morphology
types while modeling span and intensity jointly within each type.
Figure~\ref{fig:architecture} summarizes the architecture.
Appendix~\ref{app:selector} gives the full specification and training settings.

\subsection{Inference: Probe--Select--Morph}
\label{sec:method-inference}

Given an input $x$, \method{} selects one action and applies it throughout generation. To reduce the overhead of this process, the
implementation reuses computation shared by the probe and the selected
execution, and shares eligible cache storage across branch replicas.
Inference proceeds in three stages.

\paragraph{Probe.}

Let $\mathcal{I}$ denote the observed layer indices and
$L_{\mathrm{probe}}=1+\max\mathcal{I}$. We run a \basecomp{} prefill
through layers $[0,L_{\mathrm{probe}})$ to construct $H$.
We retain caches and full-prompt hidden states at reusable boundaries,
including the probe endpoint and candidate \rep{} span exits reached
during the pass.

\paragraph{Select.}

A selector predicts its highest-probability action from the complete
distribution in Equation~\ref{eq:action-distribution}.
Selection depends only on $H$ and requires no candidate response
generation.

\paragraph{Morph.}

Execution proceeds according to the selected morphology:
\begin{itemize}
\item \textbf{\basecomp{}.}
Resume from the probe endpoint and execute the remaining layers.
Probe and morph together perform one ordinary prompt prefill.

\item \textbf{\rep{}.}
For span $(i,j)$, reuse the shared prefix through
$\min(L_{\mathrm{probe}},j)$, completing the first traversal of the
span if necessary. Execute $r$ additional traversals, then continue
through layers $[j,N)$. Each repeated layer occurrence maintains
its own cache history.

\item \textbf{\branch{}.}
Run a separate prefill with $K$ replicas, branching hidden states
within the span and averaging them at its exit. We retain the reference
batch shape throughout execution to preserve its numerical behavior.
Outside the span, eligible standard-attention layers share identical
cache rows across replicas; branch-specific histories remain separate.
\end{itemize}

Prefix reuse avoids recomputing the shared portion of the prompt, while
cache sharing reduces storage for replicated histories. The detailed inference optimization and cost measurement are provided in Appendix~\ref{app:speed}.

%% file: Sections/experiment.tex
\section{Experiments}
\label{experiments}

We evaluate \method{} on three frozen backbones,including Qwen3.5-27B ($N{=}64$),
Qwen3.5-4B ($N{=}32$), and Gemma4-12B ($N{=}48$), where $N$ denotes the number of layers. Each backbone uses a 25-action space consisting of
\basecomp{}, 12 \rep{} and 12 \branch{} actions, where the 
spans are selected from the screening procedure in
Appendix~\ref{app:scan}.
Appendix~\ref{app:oracle} describes the data splits, prompt protocols,
and scoring rules. Appendix~\ref{app:selector} gives the selector
architecture and training details. Appendix~\ref{app:morphology-analysis} presents the analysis of trajectory geometry and per-task dissociation plots.

We compare five strategies: \textbf{Base} (LLM itself),
\textbf{Repeat} (the best \rep{} action in the training
set), \textbf{Branch} (the best \branch{} action in the training
set),
\textbf{Bench Prior} (the best action in the training
set of each specific benchmark), and \textbf{\method{}}.

The test set contains five in-distribution benchmarks (BBH \citep{suzgun2023challenging},
ARC-Challenge \citep{clark2018think}, OpenBookQA \citep{mihaylov2018can}, GSM8K \citep{cobbe2021training}, MATH-500 \citep{lightman2024let}), and four
unseen benchmarks (2WikiMultiHop \citep{ho2020constructing}, HellaSwag \citep{zellers2019hellaswag}, TruthfulQA-MC1 \citep{lin2022truthfulqa},
GPQA-Diamond \citep{rein2023gpqa}). All reported numbers are computed on these sets, which are held out from both training and validation.
Each benchmark uses its own scoring rule (exact-match accuracy for
multiple-choice and arithmetic tasks, token-level F1 for extractive QA).
The mean score metric is computed by averaging
per-sample scores across test samples.
For the main results in Table~\ref{tab:main}, all strategies are allowed to generate their complete responses without length limit, and scoring is performed on the completed responses. We separately examine a 32-token generation limit in Table~\ref{tab:truncation} to evaluate early-answer performance: whether a correct answer is produced within this window. Under this protocol, a response that has not yet stated its answer receives no credit, even if its continuation would provide the correct answer.

\subsection{Main Results}
\label{sec:exp-results}

\begin{table}[t]
\centering
\caption{Scores on five benchmarks, evaluated on completed responses. Bold marks the highest score per row. Gain is relative improvement of \method{} over Base.}
\vspace{-0.5em}
\label{tab:main}
\small
\begin{tabular}{llcccccc}
\toprule
Backbone & Benchmark & Base & Repeat & Branch & Bench Prior & \textbf{\method{}} & Gain \\
\midrule
\multirow{5}{*}{\small Qwen3.5-27B}
  & BBH       & 0.5105 & 0.6632 & 0.4474 & 0.6632 & \textbf{0.6895} & +35.06\% \\
  & GSM8K               & 0.4120 & 0.4280 & 0.2720 & 0.4120 & \textbf{0.4360} & +5.83\% \\
  & MATH-500            & 0.5400 & 0.4800 & 0.4600 & 0.5300 & \textbf{0.5600} & +3.70\% \\
  & ARC-Challenge    & 0.6200 & 0.7667 & 0.6733 & 0.7733 & \textbf{0.7800} & +25.81\% \\
  & OpenBookQA& 0.8267 & 0.8333 & 0.7400 & 0.8333 & \textbf{0.8400} & +1.61\% \\
\midrule
\multirow{5}{*}{\small Qwen3.5-4B}
  & BBH       & 0.4105 & 0.4421 & 0.3789 & 0.4526 & \textbf{0.5211} & +26.94\% \\
  & GSM8K     & 0.1720 & 0.1600 & \textbf{0.1800} & 0.1720 & \textbf{0.1800} & +4.65\% \\
  & MATH-500  & 0.3300 & 0.2800 & 0.3000 & 0.3100 & \textbf{0.3400} & +3.03\% \\
  & ARC-Challenge    & 0.7867 & 0.7733 & 0.7467 & 0.7467 & \textbf{0.8000} & +1.69\% \\
  & OpenBookQA& 0.7867 & \textbf{0.7933} & 0.7400 & 0.7333 & \textbf{0.7933} & +0.84\% \\
\midrule
\multirow{5}{*}{\small Gemma4-12B}
  & BBH       & 0.5947 & 0.6000 & 0.5789 & 0.5947 & \textbf{0.6105} & +2.66\% \\
  & GSM8K     & 0.2320 & 0.1920 & 0.2240 & 0.2320 & \textbf{0.2360} & +1.72\% \\
  & MATH-500  & 0.2300 & 0.2000 & 0.2300 & 0.2300 & \textbf{0.2400} & +4.35\% \\
  & ARC-Challenge    & 0.7733 & 0.7667 & \textbf{0.7800} & 0.7667 & 0.7733 & +0.00\% \\
  & OpenBookQA& 0.8800 & 0.8867 & 0.8867 & 0.8800 & \textbf{0.9067} & +3.03\% \\
\bottomrule
\end{tabular}
\end{table}

Table~\ref{tab:main} compares \method{} with the unmodified backbone and fixed morphology strategies. \method{} improves over Base in most evaluated settings and outperforms Bench Prior in every backbone–benchmark pair. Because Bench Prior selects one action per benchmark using the training set, this comparison supports the value of input-dependent selection beyond a benchmark-level fixed strategy. The gains are uneven across backbones and tasks, with the largest improvements on BBH and ARC-Challenge for Qwen3.5-27B and smaller changes on Gemma4-12B.

\paragraph{Response behavior under a short output budget.}
We also observe an interesting phenomenon in Qwen3.5-27B. Despite the direct-answer prompts, the unmodified model often produces explanatory text before stating its choice. Under morphology intervention, particularly \rep{}, responses tend to state the answer more directly. This suggests that changing internal execution can affect the timing of answer production as well as task performance.
Table~\ref{tab:truncation} shows that \method{} achieves higher scores when a correct answer must be extracted in a short window, providing an additional view.

\begin{table}[t]
\centering
\caption{Short-budget scores on Qwen3.5-27B across three benchmarks. Generation is limited to 32 tokens, so a correct answer must be produced and extracted within this window. This supplementary evaluation characterizes performance under an early-answer constraint.}
\vspace{-0.5em}
\label{tab:truncation}
\small
\begin{tabular}{lcccccc}
\toprule
Benchmark & Base & Repeat & Branch & Bench Prior & \textbf{\method{}} & Gain \\
\midrule
BBH        & 0.2000 & 0.5105 & 0.1684 & 0.5105 & \textbf{0.6053} & +202.65\% \\
ARC-Challenge     & 0.3667 & 0.7333 & 0.5000 & 0.7333 & \textbf{0.7467} & +103.63\% \\
OpenBookQA & 0.5400 & 0.7867 & 0.6267 & 0.7867 & \textbf{0.7933} & +46.91\% \\
\bottomrule
\end{tabular}
\end{table}

\subsection{Generalization to Unseen Benchmarks}
\label{sec:exp-unseen}

\begin{table}[t]
\centering
\caption{Scores on four unseen benchmarks.
Bold marks the highest score per row.}
\vspace{-0.5em}
\label{tab:unseen}
\small
\begin{tabular}{llccccc}
\toprule
Backbone & Benchmark & Base & Repeat & Branch & \textbf{\method{}} & Gain \\
\midrule
\multirow{4}{*}{\small Qwen3.5-27B}
  & 2WikiMultiHop  & 0.7463 & 0.7483 & 0.6624 & \textbf{0.7708} & +3.28\% \\
  & HellaSwag      & 0.9100 & \textbf{0.9500} & 0.8700 & 0.9300 & +2.20\% \\
  & TruthfulQA-MC1 & 0.7900 & \textbf{0.8000} & 0.7300 & \textbf{0.8000} & +1.27\% \\
  & GPQA-Diamond   & 0.5800 & 0.5500 & 0.4000 & \textbf{0.6000} & +3.45\% \\
\midrule
\multirow{4}{*}{\small Qwen3.5-4B}
  & 2WikiMultiHop  & 0.5643 & 0.5527 & 0.5757 & \textbf{0.5833} & +3.37\% \\
  & HellaSwag      & 0.8500 & 0.8800 & 0.8500 & \textbf{0.8900} & +4.71\% \\
  & TruthfulQA-MC1 & \textbf{0.6700} & 0.6500 & 0.6600 & \textbf{0.6700} & +0.00\% \\
  & GPQA-Diamond   & 0.4400 & 0.4600 & 0.3800 & \textbf{0.4700} & +6.82\% \\
\midrule
\multirow{4}{*}{\small Gemma4-12B}
  & 2WikiMultiHop  & 0.1381 & 0.1606 & 0.1500 & \textbf{0.1639} & +18.68\% \\
  & HellaSwag      & 0.4300 & 0.3900 & 0.4500 & \textbf{0.4600} & +6.98\% \\
  & TruthfulQA-MC1 & \textbf{0.5200} & \textbf{0.5200} & 0.5100 & 0.4800 & -7.69\% \\
  & GPQA-Diamond   & 0.2900 & 0.2600 & 0.3100 & \textbf{0.3200} & +10.34\% \\
\bottomrule
\end{tabular}
\end{table}


\begin{table}[t]
\centering
\small
\begin{minipage}[t]{0.6\linewidth}
\centering
\captionof{table}{Action space size ablation. Nested pools (3 $\subset$ 9 $\subset$ 25) and two 13-action pools are compared.
``Oracle'' is the per-sample upper bound achievable within that pool.}
\vspace{-0.5em}
\label{tab:ablation-anchor}
\begin{tabular}{llcc}
\toprule
Backbone & Pool & Mean Score & Oracle \\
\midrule
\multirow{5}{*}{\small Qwen3.5-27B}
  & 3 actions       & 0.5786 & 0.6381 \\
  & 9 actions       & 0.5869 & 0.7000 \\
  & 13 actions (re) & 0.5964 & 0.7202 \\
  & 13 actions (br) & 0.4524 & 0.6321 \\
  & 25 actions      & 0.6083 & 0.7619 \\
\midrule
\multirow{5}{*}{\small Qwen3.5-4B}
  & 3 actions       & 0.4738 & 0.5286 \\
  & 9 actions       & 0.4833 & 0.5571 \\
  & 13 actions (re) & 0.4810 & 0.5524 \\
  & 13 actions (br) & 0.4786 & 0.5607 \\
  & 25 actions      & 0.4964 & 0.6107 \\
\midrule
\multirow{5}{*}{\small Gemma4-12B}
  & 3 actions       & 0.5310 & 0.5583 \\
  & 9 actions       & 0.5369 & 0.6440 \\
  & 13 actions (re) & 0.5321 & 0.6750 \\
  & 13 actions (br) & 0.5274 & 0.6583 \\
  & 25 actions      & 0.5369 & 0.7071 \\
\bottomrule
\end{tabular}
\end{minipage}\hfill
\begin{minipage}[t]{0.38\linewidth}
\centering
\captionof{table}{Selector input ablation. ``Hidden trajectory ($n$)'' denotes a hidden trajectory composed of $n$ layer.}
\vspace{-0.5em}
\label{tab:ablation-input}
\begin{tabular}{lc}
\toprule
Input & Mean Score \\
\midrule
Prompt embedding        & 0.5714 \\
Single hidden state     & 0.5940 \\
Hidden trajectory (3)   & 0.5964 \\
Hidden trajectory (5)   & 0.5952 \\
Hidden trajectory (10)  & 0.6083 \\
\midrule
Prompt embedding        & 0.4643 \\
Single hidden state     & 0.4845 \\
Hidden trajectory (3)   & 0.4905 \\
Hidden trajectory (5)   & 0.4952 \\
Hidden trajectory (10)  & 0.4964 \\
\midrule
Prompt embedding        & 0.5321 \\
Single hidden state     & 0.5286 \\
Hidden trajectory (3)   & 0.5333 \\
Hidden trajectory (5)   & 0.5345 \\
Hidden trajectory (10)  & 0.5369 \\
\bottomrule
\end{tabular}
\end{minipage}
\end{table}

Table~\ref{tab:unseen} evaluates each backbone’s selector on four benchmarks excluded from selector training, without retraining or refitting. SelfMorph improves over Base in most settings, indicating that the learned selection policy can transfer beyond its training benchmarks.

\subsection{Ablation Studies}
\label{sec:exp-ablations}

We examine two central design choices: the composition of the action
space and the representation used for selection. By default, all ablations are conducted on three backbones and short-budget protocol. All remaining hyperparameters kept fixed unless otherwise noted. The mean score metric denotes the per‑sample average score. Additional
ablations of selector architecture, capacity, target temperature,
and cost coefficient are provided in
Appendix~\ref{app:additional-ablations}.

\paragraph{Action space composition.}
We train selectors on three nested pools: 3 actions (\basecomp{} and
one action per intervention family), 9 actions (\basecomp{} and four
spans per family at one intensity), and the full 25 actions (three
intensities per span). For the 3-action pool, the span for each
intervention family is selected by mean training-set utility.
We also evaluate two 13-action pools, each containing \basecomp{}
and all 12 actions from either \rep{} or \branch{}.
Table~\ref{tab:ablation-anchor} reports selector performance and the
per-sample oracle ceiling for each pool. Expanding the mixed pool
improves performance on both Qwen backbones, whereas Gemma matches
its 9-action score with 25 actions. The 9-action mixed pool
outperforms the larger single-family pools in most comparisons,
supporting the value of access to both morphology families over
additional intensity choices within one family. However, the
oracle--selector gap grows as the mixed pool expands: additional
actions increase the available headroom faster than the selector
captures it.

\paragraph{Selector input.}
We compare prompt embeddings, a single intermediate-layer hidden
state, uniformly subsampled 3- and 5-layer trajectories, and the
default 10-layer trajectory (Table~\ref{tab:ablation-input}).
Almost every hidden state condition outperforms
the prompt embedding, suggesting that the model’s intermediate computation provides stronger signal about morphology than the surface text.
The full trajectory achieves the highest score among the tested
representations on all three backbones. This supports using multiple
depths of the backbone's prompt computation to inform morphology
selection.

\vspace{-0.5em}
\subsection{Analysis}
\label{sec:exp-analysis}
\vspace{-0.5em}

\begin{figure}[t]
\centering
\includegraphics[width=\textwidth]{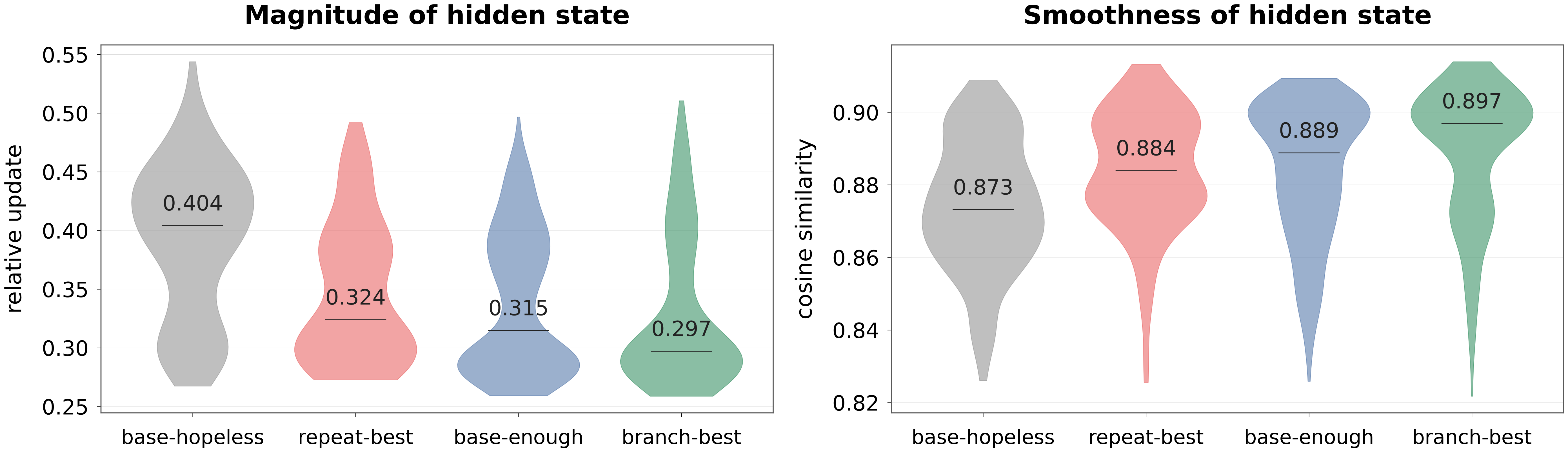}
\vspace{-0.5em}
\caption{The trajectory features before morphing. Four groups are
defined by which family the per-sample oracle selects: 
base-hopeless (oracle is \basecomp{} but sample is unsolved), 
repeat-best, base-enough (\basecomp{} and already correct), branch-best. (a)~Relative hidden-state update near the readout point.
(b)~Minimum cosine similarity across the trajectory window.}
\label{fig:regime}
\end{figure}

\paragraph{Trajectory features and morphology preference.}
To visualize the relationship between hidden trajectory patterns and preferred morphology, we examine whether trajectories differ between samples whose oracle actions belong to distinct families. Figure~\ref{fig:regime} partitions samples into four categories: \emph{base‑hopeless} (all actions defeat), \emph{base‑enough} (\basecomp{} has the highest utility), \emph{repeat‑best}, and \emph{branch‑best}. Two features are
computed from the trajectory window, before any
morphology is applied: (a)~the relative hidden state update between the
last two observed layers, and (b)~the minimum cosine similarity of adjacent observed layers across the window.

The group summaries exhibit a consistent ordering. Relative updates decrease from base‑hopeless through repeat‑best and base‑enough to branch‑best, while cosine similarity increases in the same order. Thus, the oracle groups differ in how strongly their representations change across the observed depth window. 
We interpret this pattern through a hypothesis about the role of additional computation. Serial reuse may help when further transformations of an evolving representation are useful; perturbation and aggregation may help when a relatively settled representation benefits from exploring nearby states. This interpretation is consistent with the greater trajectory activity of repeat‑best relative to branch‑best. It motivates a distinction between computational stability and computational sufficiency: the branch‑best group has relatively stable trajectories despite benefiting from intervention, while the most active group remains unsolved under the evaluated protocol.

\begin{figure}[t]
\centering
\includegraphics[width=0.95\textwidth]{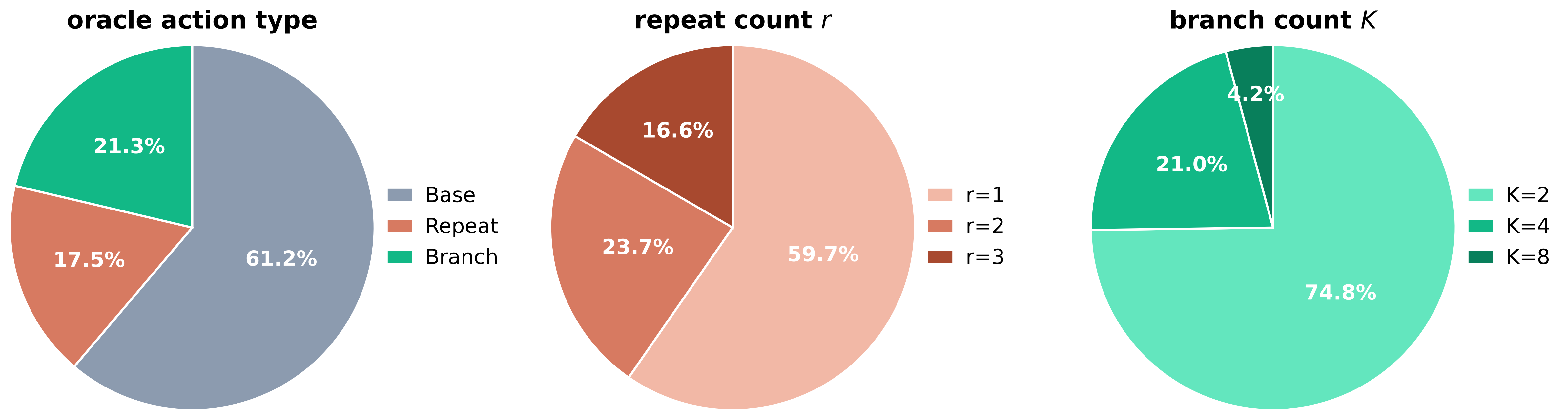}
\vspace{-0.5em}
\caption{Oracle action distribution. Left: morphology type breakdown. Center:
intensity distribution within \rep{}. Right: intensity distribution
within \branch{}.}
\label{fig:oracle-dist}
\end{figure}

\paragraph{Oracle action distribution.}
Figure~\ref{fig:oracle-dist} shows the oracle-optimal action
distribution on Qwen3.5-27B. The majority of samples are best
served by \basecomp{}. This group contains both inputs that the backbone already answers correctly
and inputs that no morphology can rescue, so \basecomp{} wins by
incurring zero extra cost.
For both \rep{} and \branch{}, the settings with the lowest intensity receive the largest share of selections. And a non‑trivial set of problems demands larger intervention intensities, as higher intensity tackle the problems where low intensity fails.

%% file: Sections/conclusion.tex
\section{Conclusion}
\label{conclusion}
We introduced \method{}, a framework that makes the internal computation of frozen LLMs an input-dependent choice over the original forward path, serial layer reuse, and local branching. Trained through utility-based distillation of offline interventions, a lightweight selector reads the backbone’s prompt trajectory to choose the morphology type, span, and intensity before generation. Across three backbones and nine benchmarks, \method{} demonstrates the value of input-dependent selection and transfers to unseen benchmarks without retraining. Ablations support the contributions of morphology diversity, depth-wise trajectory information, and the architecture components. These findings establish computational shape as a useful dimension of inference-time adaptation.

%% file: Sections/appendix.tex
\input{Sections/background}

\section{Span Screening}
\label{app:scan}

Before fixing the action space $\mathcal{A}$
(Section~\ref{sec:method-space}), we screen candidate spans at a
fixed intensity for each intervention type to identify favorable
depth regions for \rep{} and \branch{}.
This screening is a diagnostic pass run on a small, fast probe set that
is separate from the formal evaluation benchmarks.

\subsection{Setup}

\paragraph{Backbones.}
We screen three frozen backbones: Qwen3.5-27B ($N{=}64$, $d_h{=}5120$),
Qwen3.5-4B ($N{=}32$, $d_h{=}2560$), and Gemma4-12B ($N{=}48$,
$d_h{=}3840$), where $N$ is the backbone depth and $d_h$ is the dimension of hidden state. Every weight parameter is in bfloat16 precision.

\paragraph{Probe set.}
Screening over thousands of span candidates requires a probe set that is
both cheap to evaluate and informative about general model capability.
Following by \cite{ng2026rys}, we use two diagnostic datasets, each containing 16 samples. The first
presents large-integer estimation problems (square roots, cube roots,
multi-digit products) scored by relative numerical error with a
digit-length correction. The second presents emotion-intensity prediction
items in an EQ-Bench format, scored by absolute deviation from reference. The two tasks are cognitively orthogonal: one tests abstract numerical intuition,
the other tests social and affective reasoning. Using both provides a broader basis for screening than either diagnostic task alone. 
Both datasets produce
short outputs (a single number or a few scores), which keeps each
generation fast enough for exhaustive enumeration.

For the numerical estimation probe, the scorer extracts all integers from
the generated text, computes for each a relative error against the
reference answer, and applies a digit-length correction that penalizes
guesses whose number of digits differs from the reference. The score for
a single extraction is $(1 - \text{relative error}) \times (1 -
\text{padding fraction})$, clamped to $[0,1]$, and the sample score is
the maximum over all extracted integers. For the emotion-intensity probe,
the scorer parses four predicted intensity values (on a 0--10 scale) from
the generated text, computes their summed absolute deviation from the
reference, and normalizes by the maximum possible deviation (40).
Inputs are formatted as raw-text continuations without a chat template.
Generation uses greedy decoding.

\paragraph{Scan grid.}
For each backbone we enumerate every half-open span $(i,j)$ with
$0 \le i < j \le N$. \rep{} is tested at $r{=}1$ and \branch{} at
$K{=}4$ with $\sigma_{\max}{=}1.0$, using the same perturbation and
hidden-space mean merge described in Section~\ref{sec:method-space}.
The screening implementation excludes $j{=}N$ for \branch{}
because its merge hook requires a layer at position $j$. The full grid contains $N(N{+}1)/2$ spans for \rep{} and
$N(N{+}1)/2 - N$ for \branch{}, yielding 4097 / 1025 / 2305
actions per sample for the 64 / 32 / 48-layer backbones. All grids were
executed to completion with zero missing entries. Scan parameters are
identical across all three backbones.

\subsection{Heatmaps}

\begin{figure}[h]
\centering
\includegraphics[width=\textwidth]{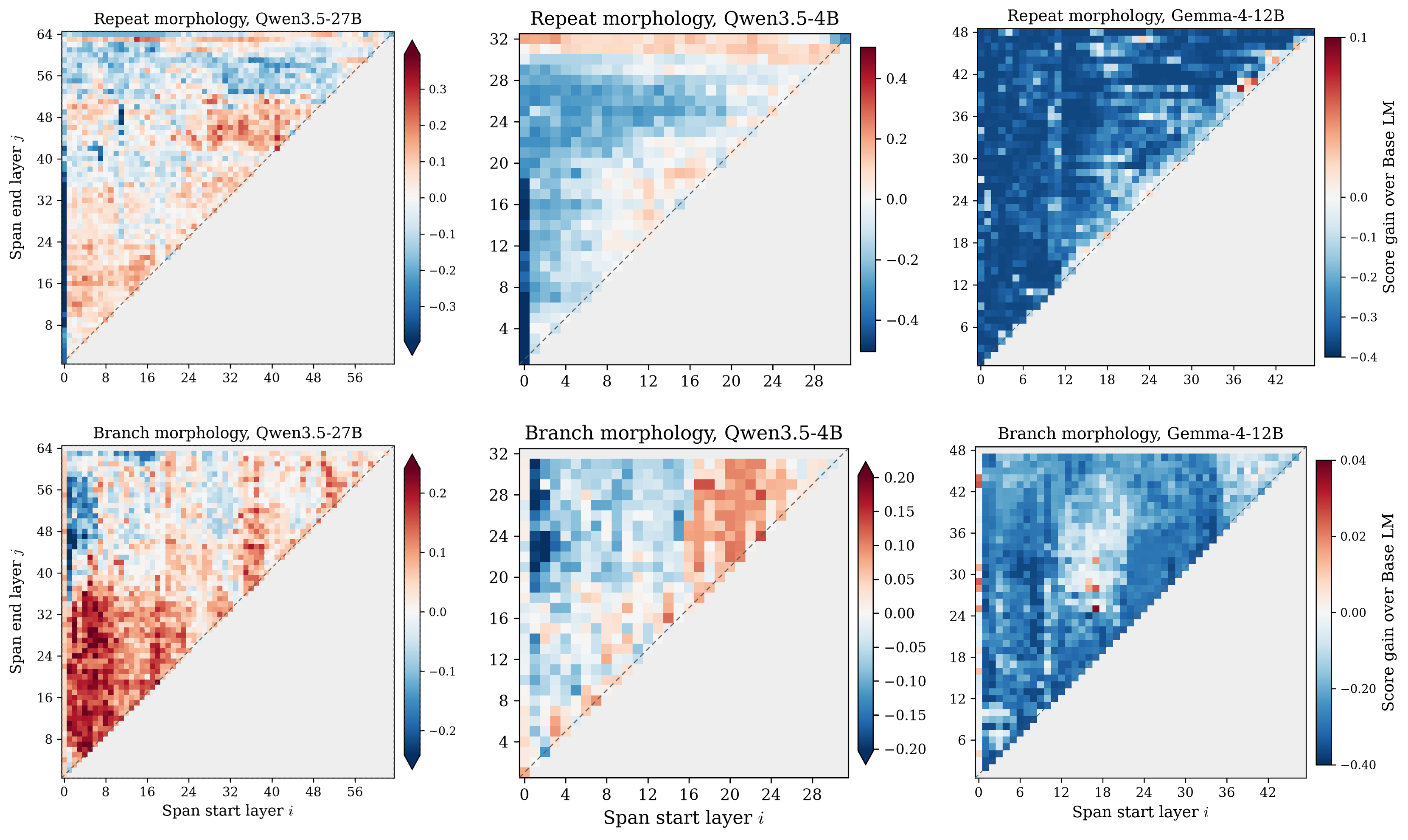}
\caption{Score gain $\Delta(i,j)$ on the numerical
estimation probe. Top: \rep{} ($r{=}1$). Bottom: \branch{} ($K{=}4$).
Columns left to right: Qwen3.5-27B ($N{=}64$), Qwen3.5-4B ($N{=}32$),
Gemma4-12B ($N{=}48$). The $x$-axis is span start $i$, the $y$-axis
is span end $j$, and only the upper triangle ($i<j$) is valid.}
\label{fig:scan-math}
\end{figure}

\begin{figure}[h]
\centering
\includegraphics[width=\textwidth]{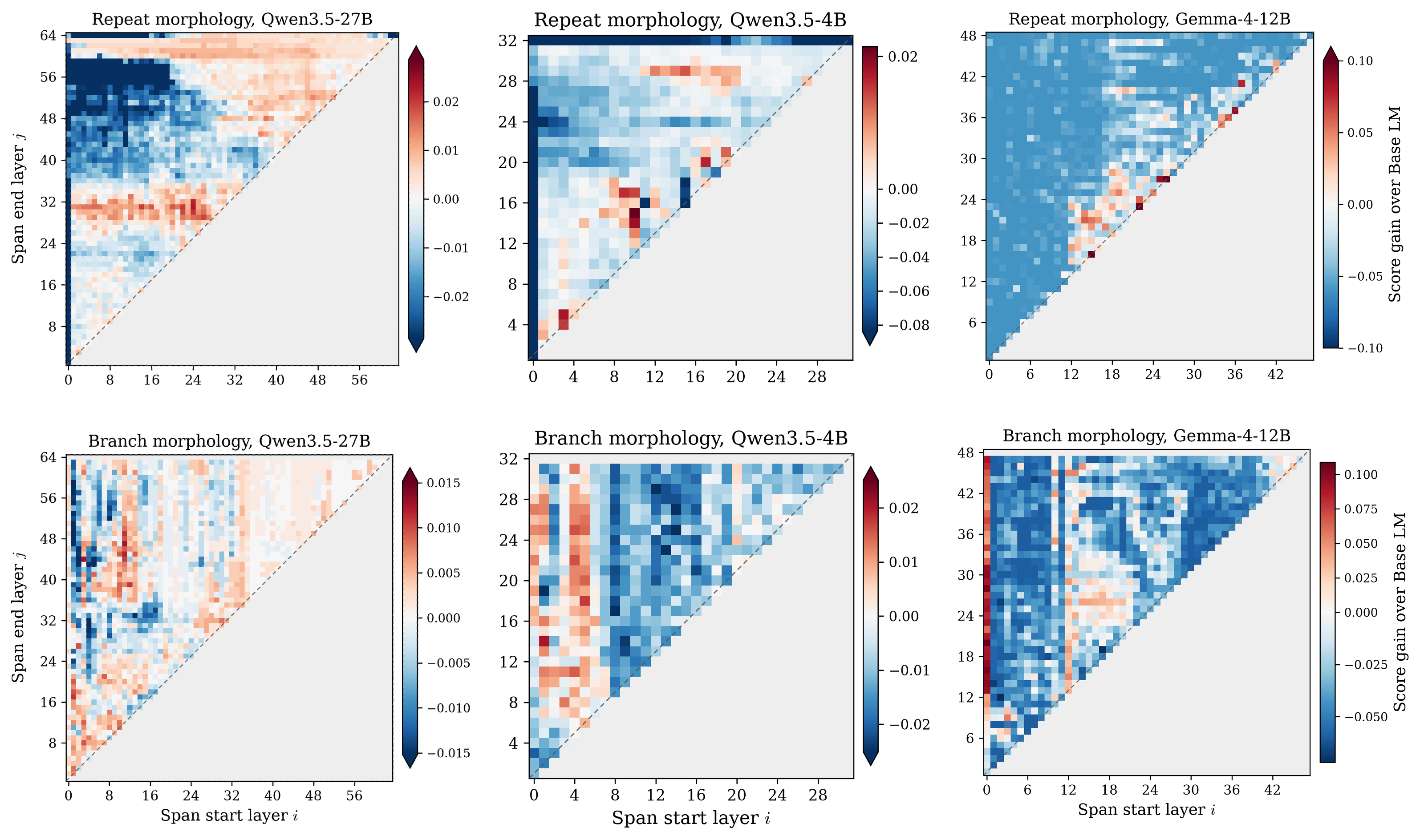}
\caption{Score gain $\Delta(i,j)$ over \basecomp{} on the
emotion-intensity probe. Top: \rep{} ($r{=}1$). Bottom: \branch{} ($K{=}4$).
Columns left to right: Qwen3.5-27B ($N{=}64$), Qwen3.5-4B ($N{=}32$),
Gemma4-12B ($N{=}48$). The $x$-axis is span start $i$, the $y$-axis
is span end $j$, and only the upper triangle ($i<j$) is valid.}
\label{fig:scan-eq}
\end{figure}

For a fixed morphology type and screening intensity, each cell
$(i,j)$ shows the mean score gain over \basecomp{} on the
16 samples in the diagnostic dataset $\mathcal{D}_{\mathrm{probe}}$:
\[
\Delta(i,j)
=\frac{1}{16}\sum_{x\in\mathcal{D}_{\mathrm{probe}}}
\left[
\mathrm{score}_{(i,j)}(x)
-\mathrm{score}_{\basecomp{}}(x)
\right].
\]
Figures~\ref{fig:scan-math}
and~\ref{fig:scan-eq} show the results, with rows corresponding to
\rep{} and \branch{} and columns to the three backbones. Red indicates
improvement and blue indicates degradation.

\subsection{Observations}

\paragraph{Intervention effects depend on span location.}
Many spans produce negative gains, including spans that begin in
the earliest layers. The affected regions vary across morphology
types, backbones, and diagnostic tasks, motivating explicit screening.

\paragraph{\rep{} gains concentrate at mid-depth.}
On all three backbones, the positive region for \rep{} occupies a band
in the mid-to-late depth range. This is the depth range where layers iteratively
refine task-relevant features \citep{tenney2019bert,lad2025remarkable},
and composing $f_s$ with itself continues to advance the representation
(Section~\ref{sec:method-space}). The pattern holds across both probe
datasets and all three model scales.

\paragraph{\branch{} gains concentrate at shallow depth.}
On Qwen3.5-27B, the positive region for \branch{} occupies the
lower-left corner of the heatmap (small $i$, moderate $j$), with 72\%
of all \branch{} spans producing a positive delta on the numerical
probe. This is consistent with the shallow layers still holding a malleable
representation \citep{tenney2019bert,lad2025remarkable}: a perturbation
introduced there can still be resolved into a genuinely different
downstream computation. On the smaller backbones
(4B and 12B), the positive region for \branch{} is broader and extends
into mid-depth spans, with the numerical-probe peak at $(23,28)$ for 4B
and $(17,25)$ for Gemma. This broader spread is consistent with the
observation that smaller models have less cleanly separated functional
stages: encoding, refinement, and decoding overlap more across depth,
so the depth at which branching pays off is less sharply localized.
The overall pattern remains that \branch{} gains start shallow and taper
off toward the output end of the stack, while \rep{} gains concentrate
in the middle.

\paragraph{The two probe datasets rank spans differently.}
The Spearman rank correlation between the numerical and emotion probes is
close to zero on the Qwen backbones ($\rho \approx -0.06$ and $+0.06$)
and moderate on Gemma ($\rho = +0.63$). A span that helps on one probe
does not reliably help on the other. We therefore rank candidates by a
balanced criterion: for each probe, the raw mean $\Delta$ of every span
is standardized to zero mean and unit variance ($z$-score), and the two
$z$-scores are summed.

\subsection{From Screening to Action Space}

This screening step produces a set of candidate options and the selector
subsequently chooses an action for each input.
Candidate spans are ranked by the balanced $z$-score (restricted to
widths $j{-}i \le 10$ to keep analytic cost moderate) and then validated
on a larger sample of the same two probe tasks, scored with the procedures described above. Each backbone selects its own spans from its own scan.

Table~\ref{tab:final-spans} lists the spans selected for each backbone.
The selected \rep{} spans lie near the middle of Qwen3.5-27B,
toward the end of Qwen3.5-4B, and across intermediate and later
depths of Gemma4-12B. The selected \branch{} spans occur early
in Qwen3.5-27B and cover a wider depth range in the smaller backbones.

\begin{table}[h]
\centering
\caption{Layer spans in the final action space $\mathcal{A}$ for each
backbone. Each span is paired with three intensity levels
($r \in \{1,2,3\}$ for \rep{}, $K \in \{2,4,8\}$ for \branch{}).}
\label{tab:final-spans}
\begin{tabular}{llccc}
\toprule
Type & & Qwen3.5-27B & Qwen3.5-4B & Gemma4-12B \\
\midrule
\multirow{4}{*}{\rep{}} & Span 1 & $(33,35)$ & $(30,31)$ & $(36,37)$ \\
& Span 2 & $(31,35)$ & $(28,31)$ & $(26,27)$ \\
& Span 3 & $(30,36)$ & $(24,31)$ & $(15,16)$ \\
& Span 4 & $(26,35)$ & $(24,30)$ & $(25,27)$ \\
\midrule
\multirow{4}{*}{\branch{}} & Span 1 & $(1,10)$ & $(23,28)$ & $(0,9)$ \\
& Span 2 & $(4,7)$ & $(3,8)$ & $(0,10)$ \\
& Span 3 & $(3,8)$ & $(3,7)$ & $(3,7)$ \\
& Span 4 & $(5,7)$ & $(22,26)$ & $(16,26)$ \\
\bottomrule
\end{tabular}
\end{table}

\section{Oracle Collection and Data Splits}
\label{app:oracle}

The selector in \method{} is trained on per-sample utility values
obtained by running every action in $\mathcal{A}$ on every training
input and recording the result. This appendix describes the datasets,
prompt protocols, scoring rules, and collection process behind those
measurements.

\subsection{Benchmark Descriptions}

We evaluate on nine benchmarks. Five
benchmarks are used for training and in-distribution evaluation, and
four additional benchmarks are held out entirely from training for
zero-shot generalization assessment.

\textbf{BBH} \citep{suzgun2023challenging} is a curated subset of
BIG-Bench tasks where language models fall below average human
performance.

\textbf{ARC-Challenge} \citep{clark2018think} contains grade-school
science questions filtered to those that simple retrieval and
co-occurrence methods answer incorrectly.

\textbf{OpenBookQA} \citep{mihaylov2018can} poses science questions that
require combining a provided fact with broader commonsense knowledge.

\textbf{GSM8K} \citep{cobbe2021training} consists of grade-school
mathematical word problems requiring multi-step arithmetic.

\textbf{MATH-500} \citep{lightman2024let} samples 500 competition-level
mathematics problems covering algebra, geometry, number theory, and
combinatorics.

\textbf{2WikiMultiHop} \citep{ho2020constructing} is a multi-hop
extractive QA task where each question requires combining information
from at least two Wikipedia paragraphs.

\textbf{HellaSwag} \citep{zellers2019hellaswag} tests commonsense
reasoning by asking the model to choose the most plausible continuation
of a sentence from four candidates.

\textbf{TruthfulQA-MC1} \citep{lin2022truthfulqa} evaluates whether
models can identify the single truthful answer among plausible
distractors. Each question has a variable number of candidates (2 to
13).

\textbf{GPQA-Diamond} \citep{rein2023gpqa} contains graduate-level
science questions (physics, chemistry, biology) written by domain
experts under an adversarial ``Google-proof'' protocol that discards
questions answerable by web search.

\subsection{Datasets and Splits}

We evaluate on five in-distribution benchmark units drawn from three
domains: ARC-Challenge and OpenBookQA (science multiple-choice), GSM8K
and MATH-500 (mathematics), and BBH (reasoning). The four unseen benchmarks  use separate frozen sample lists and are evaluated with the
production selector without any refitting.

Table~\ref{tab:splits} reports the in-distribution sample counts. The
three splits serve distinct roles: the training set supplies the oracle
utilities for selector training, the validation set (the union of two
equal-sized sub-pools) is used exclusively for early stopping within a
training run, and the test set is held out from both training and early
stopping. The test set is never seen during any part of the training
process, and the numbers reported in Section~\ref{experiments} are
computed on it. Samples are drawn by stratified sampling over subtasks or
dataset splits with a fixed random seed \texttt{20260810}. All three backbones
use the same sample manifest.

\begin{table}[h]
\centering
\caption{Sample counts per benchmark and split. All three backbones use
the same samples, drawn from a single frozen sample manifest with a fixed
random seed.}
\label{tab:splits}
\begin{tabular}{lcccc}
\toprule
Benchmark & Train & Validation & Test & Total \\
\midrule
BBH & 2619 & 266 & 190 & 3075 \\
ARC-Challenge & 779 & 200 & 150 & 1129 \\
OpenBookQA & 623 & 200 & 150 & 973 \\
GSM8K & 2922 & 500 & 250 & 3672 \\
MATH500 & 350 & 50 & 100 & 500 \\
\midrule
\textbf{Total} & \textbf{7293} & \textbf{1216} & \textbf{840} & \textbf{9349} \\
\bottomrule
\end{tabular}
\end{table}

\subsection{Prompt Templates}

Below we describe the prompt templates used for each benchmark family. Demonstrations are fixed across all samples within a benchmark. All five training benchmarks use direct-answer prompts to encourage concise responses during exhaustive oracle collection. The short-budget protocol is used for oracle collection and the ablations described in Section~\ref{experiments}. For the main results in Table~\ref{tab:main}, generation proceeds until the model completes its response, without an imposed generation-length limit. Thus, the direct-answer instruction specifies the requested response style; the generation limit is a separate evaluation setting.

\paragraph{ARC-Challenge and OpenBookQA.}
Three fixed demonstrations precede the test question. Each demonstration
and the test item follow the format:

{\small
\begin{verbatim}
Q: <question text>
Options:
(A) <option 1>
(B) <option 2>
(C) <option 3>
(D) <option 4>
A: (<letter>)
\end{verbatim}
}

\noindent The model generates from \verb|A: | onward. ARC-Challenge and
OpenBookQA each use their own set of three demonstrations.

\paragraph{GSM8K.}
A system instruction requests that the model output only the final
number without showing any explicit reasoning. Two demonstrations follow the
instruction, each ending with the bare numerical answer. The test item
is formatted as:

{\small
\begin{verbatim}
<instruction: output only the final number>

Q: <demo 1>
A: The answer (arabic numerals) is <number>

Q: <demo 2>
A: The answer (arabic numerals) is <number>

Q: <test question>
A: The answer (arabic numerals) is
\end{verbatim}
}

\noindent The scorer extracts the
last number from the output and compares it numerically to the reference
($|\text{pred} - \text{gold}| < 10^{-3}$).

\paragraph{MATH500.}
The same structure as GSM8K, with the instruction requesting a
\verb|\boxed{}| answer. Two demonstrations include boxed answers
(e.g., \verb|\boxed{2}|, \verb|\boxed{\frac{1}{2}}|). The scorer
extracts the last \verb|\boxed{}| expression and normalizes it using
standard rules (fraction simplification, percentage conversion,
equation right-hand-side extraction) before string comparison.

\paragraph{BBH.}
Each subtask uses its official BIG-Bench-Hard task description and three
demonstration questions. The demonstrations are derived from the official
chain-of-thought prompts by stripping the reasoning text and retaining
only the final answer letter:

{\small
\begin{verbatim}
<official task description>

Q: <demo question 1>
A: (<letter>)

Q: <demo question 2>
A: (<letter>)

Q: <demo question 3>
A: (<letter>)

Q: <test question>
A:
\end{verbatim}
}

\noindent The scorer extracts the
first answer letter or parses the output
for affirmative or negative keywords using a priority cascade:
parenthesized single letter, ``the answer is'' phrase, then earliest
keyword match..

\paragraph{HellaSwag, TruthfulQA-MC1, GPQA-Diamond.}
All use 0-shot multiple-choice with a direct-answer instruction
and no demonstrations:

{\small
\begin{verbatim}
Q: <question or sentence prefix>
Options:
(A) <candidate 1>
...
Answer with only the option letter.
A:
\end{verbatim}
}

\noindent Option order is deterministically shuffled per item to remove
positional bias. TruthfulQA-MC1 has a variable number of candidates (2
to 13).

\paragraph{2WikiMultiHop (extractive QA).}
The model receives a set of Wikipedia paragraphs and a multi-hop
question, and must produce a answer:

{\small
\begin{verbatim}
<instruction: answer with a short phrase>

Context:
[Title 1] <paragraph text>
[Title 2] <paragraph text>
...

Question: <multi-hop question>
Answer:
\end{verbatim}
}

\noindent The scorer first attempts
exact match (after lowercasing and stripping articles/punctuation), then
falls back to token-level F1.

\subsection{Collection Process}

For each of the 9349 samples and each of the 25 actions in
$\mathcal{A}$, we execute a single greedy generation on the frozen
backbone and record the sample identifier, action identifier, binary
score, analytic cost, utility, and the generated text. This produces
233{,}725 records per backbone.

During the \basecomp{} forward pass of each sample, we additionally
record the depth-wise hidden trajectory used as the selector's input
feature. At $L{=}10$ pre-selected layers (spaced to cover the mid-to-late
portion of each backbone), we store the last-token hidden state, yielding
a tensor of shape $[10, d_h]$ per sample. The trajectory is collected
once per sample, during the \basecomp{} pass only. Non-\basecomp{}
actions do not record trajectories, because the selector conditions on
the model's computation before any morphology is applied.

Collection is parallelized across GPUs by sharding over samples (8
shards for the 27B backbone, 16 for the 4B and 12B backbones). Each
record is flushed to disk immediately, and collection supports
interruption and exact resumption at the granularity of individual
(sample, action) pairs.

\subsection{Baselines}

Four fixed baselines are fitted on the train split per backbone. The fitting criterion in every case is the
action with the highest mean utility on the relevant subset of training
samples.

\begin{itemize}
\item \textbf{Base LM}: always selects \basecomp{}.
\item \textbf{Fixed-\rep{}}: the single \rep{} action with the highest
mean utility across all training samples.
\item \textbf{Fixed-\branch{}}: the single \branch{} action with the
highest mean utility across all training samples.
\item \textbf{Bench Prior}: one action per benchmark unit, each
chosen as the highest-mean-utility action on that unit's training
samples. Different units may select different actions, but all samples
within a unit share the same action.
\end{itemize}

\section{Selector Architecture and Training}
\label{app:selector}

This appendix specifies the selector network and training procedure for
each backbone. The Method section describes the selector functionally
(Section~\ref{sec:method-selector}). Here we give the concrete dimensions,
parameter counts, and optimization settings.

\subsection{Input Representation}

The selector receives a depth-wise hidden trajectory
$H \in \mathbb{R}^{L \times d_h}$ extracted during the \basecomp{}
forward pass, with $L{=}10$ for
all three backbones. The 10 layers are chosen to span the mid-to-late
portion of each backbone: layers 20, 22, \ldots, 38 for Qwen3.5-27B,
layers 12, 14, \ldots, 30 for Qwen3.5-4B, and layers 16, 18, \ldots, 34 for Gemma4-12B.

Before entering the network, each layer slice $H_l$ is passed through
RMSNorm ($\epsilon{=}10^{-6}$) and scaled elementwise by $\gamma_l$, the
pretrained gain vector of the backbone's own input layer normalization at
the consuming layer. These gains are frozen buffers, not trainable
parameters. The normalization aligns the selector's input with the
numerical semantics the backbone uses at each depth.

\subsection{Network Architecture}

\begin{figure}[h]
\centering
\includegraphics[width=\textwidth]{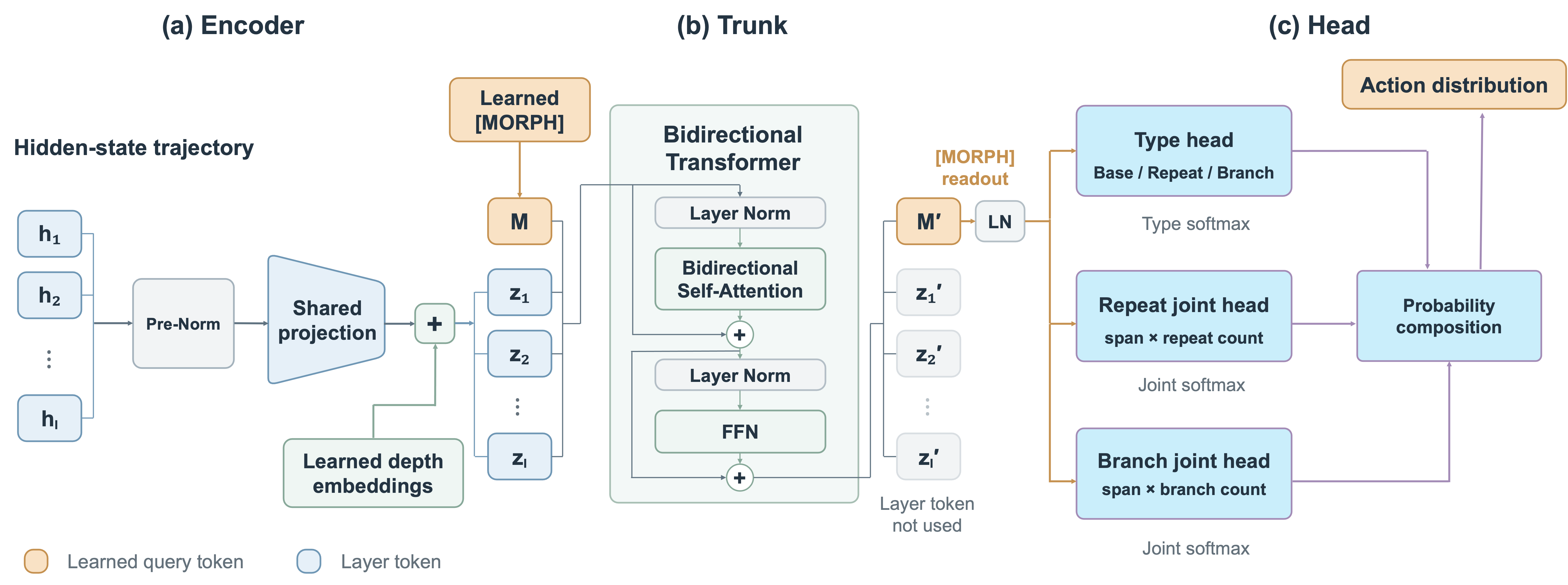}
\caption{Selector architecture. (a)~Encoder: per-layer hidden states are
pre-normalized and projected to a shared low-dimensional space, then
combined with learned depth embeddings. (b)~Trunk: a learned query token
$[\mathrm{MORPH}]$ and the $L$ layer tokens attend to one another in a
bidirectional transformer block. The output at the query position serves
as the decision vector. (c)~Head: the decision vector is mapped to a
type distribution and two joint span-intensity distributions, composed
into the final action distribution.}
\label{fig:architecture}
\end{figure}

Figure~\ref{fig:architecture} illustrates the selector's three-stage
architecture. We describe each stage below with the computation it
performs.

\paragraph{(a) Encoder.}
Each layer slice $H_l \in \mathbb{R}^{d_h}$ first passes through the
pre-normalization described in the preceding subsection. A single shared
linear projection maps the result to a $d$-dimensional feature:
\[
s_l = W_{\mathrm{proj}}\,\mathrm{PreNorm}(H_l), \qquad
W_{\mathrm{proj}} \in \mathbb{R}^{d \times d_h}, \quad l = 1,\dots,L.
\]
On Gemma4-12B, the encoder additionally computes a first-order
difference $\delta_l = s_l - s_{l-1}$ (with $\delta_1=\mathbf{0}$)
and concatenates it with $s_l$, producing a $2d$-dimensional feature per
layer.

A fusion layer projects each feature to the trunk width $d_z$ and adds a
learned depth embedding:
\[
z_l = W_{\mathrm{fuse}}\,s_l + e_l, \qquad
W_{\mathrm{fuse}} \in \mathbb{R}^{d_z \times d}, \quad
e_l \in \mathbb{R}^{d_z},
\]
where $\{e_1,\dots,e_L\}$ is a learned embedding table of size
$L \times d_z$, the only source of positional information in the network.

\paragraph{(b) Trunk.}
A learned query token $\mathbf{M} = [\mathrm{MORPH}] \in \mathbb{R}^{d_z}$
is prepended to the sequence of $L$ layer tokens, forming a
length-$(L{+}1)$ input:
\[
\mathbf{Z}^{(0)} = [\,\mathbf{M},\; z_1,\; z_2,\; \dots,\; z_L\,]
\in \mathbb{R}^{(L+1) \times d_z}.
\]
This sequence is processed by a single pre-norm bidirectional
transformer block with $n_h$ attention heads and an FFN of expansion
factor $f{=}4$. All $L{+}1$ tokens attend to one another without a
causal mask:
\begin{align}
&\hat{\mathbf{Z}} = \mathbf{Z}^{(0)}
  + \mathrm{BiAttn}\!\big(\mathrm{LN}(\mathbf{Z}^{(0)})\big), \\
&\mathbf{Z}^{(1)} = \hat{\mathbf{Z}}
  + \mathrm{FFN}\!\big(\mathrm{LN}(\hat{\mathbf{Z}})\big).
\end{align}
The output at position 0 (the query token), passed through a final
LayerNorm, serves as the decision vector:
\[
z_{\mathrm{out}} = \mathrm{LN}\!\big(\mathbf{Z}^{(1)}_0\big)
\in \mathbb{R}^{d_z}.
\]
The remaining $L$ layer-token outputs are discarded. Information from
each depth reaches $z_{\mathrm{out}}$ through bidirectional attention,
which acts as a learned, content-adaptive pooling over the trajectory.

\paragraph{(c) Head.}
The head maps $z_{\mathrm{out}}$ to a distribution over $\mathcal{A}$
through three parallel branches.

\emph{Type head.} A linear layer produces logits over
$\{\basecomp{}, \rep{}, \branch{}\}$:
\[
\ell_\tau = W_\tau\, z_{\mathrm{out}} + b_\tau \in \mathbb{R}^3,
\qquad p_\phi(\tau \mid H) = \mathrm{softmax}(\ell_\tau).
\]

\emph{Repeat joint head.} A two-layer MLP produces $S \times P$ joint
logits over span and intensity:
\[
\ell_{\rep} = W_2^{\rep}\,\mathrm{GELU}\!\big(W_1^{\rep}\,z_{\mathrm{out}}\big)
\in \mathbb{R}^{S \cdot P},
\]
reshaped to $[S, P]$ and passed through a single softmax over all
$S{\cdot}P$ entries:
$p_\phi(s, \theta \mid \rep, H)
= \mathrm{softmax}(\ell_{\rep})_{s,\theta}$.

\emph{Branch joint head.} Same structure as \textit{Repeat joint head}.

\emph{Probability composition.} The final action probability is:
\[
p_\phi(m\mid H)=
\begin{cases}
p_\phi(\basecomp{}\mid H), & m=\basecomp{},\\
p_\phi(\tau\mid H)\,
p_\phi(s,\theta\mid\tau,H), & m=(\tau,s,\theta).
\end{cases}
\]
For all three backbones $S{=}4$ spans and $P{=}3$ intensity levels.

\subsection{Dimensions and Parameter Counts}

Table~\ref{tab:selector-dims} lists the concrete dimensions. The
selector is small relative to the backbone.

\begin{table}[h]
\centering
\caption{Selector dimensions and parameter counts. All parameters are
trainable. The pre-norm gains ($\gamma_l$) are frozen buffers and not
included.}
\label{tab:selector-dims}
\begin{tabular}{lrrr}
\toprule
& Qwen3.5-27B & Qwen3.5-4B & Gemma4-12B \\
\midrule
Backbone $d_h$ & 5120 & 2560 & 3840 \\
Encoder output $d$ & 192 & 192 & 384 \\
Trunk input & $d$ & $d$ & $2d$ \\
Trunk width $d_z$ & 64 & 64 & 128 \\
Attention heads $n_h$ & 2 & 2 & 4 \\
Blocks $n_{\mathrm{blk}}$ & 1 & 1 & 1 \\
Head hidden $d_{\mathrm{head}}$ & 48 & 48 & 96 \\
\midrule
Encoder params & 983{,}232 & 491{,}712 & 1{,}474{,}944 \\
Trunk params & 63{,}168 & 63{,}168 & 298{,}368 \\
Head params & 7{,}611 & 7{,}611 & 27{,}483 \\
\midrule
\textbf{Total per selector} & \textbf{1{,}054{,}011} & \textbf{562{,}491} & \textbf{1{,}800{,}795} \\
\bottomrule
\end{tabular}
\end{table}

\subsection{Training}

Table~\ref{tab:training} lists the optimization settings. Training
minimizes the distillation loss in Equation~\ref{eq:distillation}
against the teacher distribution $q_T(m\mid x)$.

\begin{table}[h]
\centering
\caption{Training hyperparameters. Values that differ across backbones
are separated by slashes (27B / 4B / Gemma).}
\label{tab:training}
\begin{tabular}{lr}
\toprule
Optimizer & AdamW \\
Peak learning rate & $8 \times 10^{-4}$ / $8 \times 10^{-4}$ / $7 \times 10^{-4}$ \\
Schedule & Cosine decay, 5\% linear warmup, min = 2\% of peak \\
Gradient clipping & Global norm 5.0 \\
Effective batch / micro-batch & 256 / 128 (accumulation = 2) \\
Max epochs & 50 / 50 / 50 \\
Early stopping patience (epochs) & 8 / 8 / 8 \\
Early stopping metric & Mean utility on validation set \\
Improvement threshold & $10^{-5}$ \\
Distillation temperature $T$ & 0.15 / 0.01 / 0.5 \\
Cost coefficient $\lambda$ & 0.05 \\
Head dropout & 0.05 / 0.05 / 0.1 \\
\bottomrule
\end{tabular}
\end{table}

Early stopping monitors mean utility on the validation split, using the highest-probability
action predicted by the selector for each input. The checkpoint
with the highest validation mean utility is retained.

\section{Inference Optimization and Cost Measurement}
\label{app:speed}

A direct implementation of \method{} evaluates overlapping prompt
prefixes during probing and execution, and stores identical cache rows
outside the branch span. We eliminate these redundancies through
prefix reuse and lossless cache deduplication, respectively. The
optimizations preserve the selected morphology, model weights,
precision, perturbations, and decoding rule. This appendix describes
their equivalence conditions and examines the resulting time and
memory costs.

\subsection{Reusing Computation Across Probe and Commit}
\label{app:speed-reuse}

\paragraph{Truncated probing and prefix reuse.}
Let $\mathcal{I}$ be the observed layer indices and
$L=1+\max\mathcal{I}$. The probe executes layers $[0,L_{\mathrm{probe}})$ and collects
the requested last-token states during prefill. Layers beyond $L_{\mathrm{probe}}-1$
cannot affect the trajectory and are omitted. We retain the full
prompt hidden states and caches at reusable boundaries.

For \basecomp{}, execution resumes at depth $L_{\mathrm{probe}}$, so probe and commit
together perform one ordinary prefill. For \rep{}, the layer sequence is
\[
    [0,j)\; +\; [i,j)\times r\; +\; [j,N).
\]
The first $j$ evaluations are shared with the ordinary forward pass.
We therefore reuse the first $\min(L_{\mathrm{probe}},j)$ evaluations, then continue
along the selected sequence. Repeated layer occurrences maintain
separate cache slots, and mutable states are copied before reuse to
protect the saved prefix. Across probe and commit, the prompt-level
layer counts are
\[
    E_{\basecomp{}}=N,\qquad
    E_{\mathrm{REP}}=N+r(j-i)+\max(L_{\mathrm{probe}}-j,0).
\]
The last term accounts for probe evaluations beyond the point where
\rep{} diverges from the ordinary path. Reuse preserves the inputs,
layer order, and cache states of the shared computation. Exact
trajectory comparisons and dedicated checks of 13 prefix-compatible
actions on two prompts confirm unchanged states and generated tokens
in our evaluation environment.

\paragraph{Preserving branch execution.}
For a single input, the reference \branch{} implementation executes
the stack at batch size $K$, while branching and merging hidden states
within the selected span. Collapsing the batch outside that span
changes execution shapes and can change floating-point accumulation
through kernel selection. With identical perturbations, this
alternative changed 58 of 144 reference outputs in a controlled
comparison. We retain the original batch shape and execute a separate
branch prefill. The next optimization removes duplicate storage while
preserving that execution shape.

\subsection{Lossless Deduplication of Branch Caches}
\label{app:speed-dedup}

\paragraph{A cache invariant.}
For standard-attention layers outside the branch span, replica cache
rows are identical after prefill: the replicas process the same inputs
before branching and the same merged states after branching. Each
subsequent decoding step appends identical keys and values at these
layers, preserving equality of the complete cache histories. We exploit
this invariant by storing one row and exposing $K$ logical rows through
a broadcast view with zero stride along the replica dimension. Layers
inside the span retain their branch-specific histories.

Updates append keys and values to the single stored row before
broadcasting the result for attention. This avoids materializing
repeated histories during concatenation. Sliding-window truncation
and cache positions follow the original update rules. Attention
receives the same logical key and value tensors as in the reference
execution. Under the evaluated backend, direct comparisons for
$K\in\{2,4,8\}$ confirm bit-identical attention outputs. End-to-end
checks also produce identical text on three prompts per configuration
for Gemma with span $(0,10)$ and Qwen3.5-4B with span $(3,8)$ at $K=4$.

\paragraph{Storage follows the distinct histories.}
Let $b_\ell$ be the single-replica cache size of layer $\ell$ at a fixed
sequence length, and let $\mathcal{D}$ contain the layers eligible for
deduplication. Define $\rho$ as the fraction of single-replica cache
bytes belonging to layers that retain separate copies:
\[
    \rho=\frac{\sum_{\ell\notin\mathcal{D}}b_\ell}
                {\sum_\ell b_\ell}.
\]
The retained cache footprint is then
\[
    B_{\mathrm{dedup}}
    =\sum_{\ell\in\mathcal{D}}b_\ell
       +K\sum_{\ell\notin\mathcal{D}}b_\ell,
    \qquad
    \frac{B_{\mathrm{dedup}}}{B_{\mathrm{single}}}
    =1+(K-1)\rho,
\]
where $B_{\mathrm{single}}=\sum_\ell b_\ell$. With uniform
standard-attention caches, only the span retains multiple histories,
so $\rho=(j-i)/N$. The byte-weighted expression also accommodates
sliding-window and hybrid architectures.

The Qwen backbones additionally maintain linear-attention recurrent
and convolutional states. Their current interfaces update fixed-address
buffers in place, which does not support writes to aliased broadcast
rows. These states remain replicated and contribute to $\rho$.
Consequently, savings depend on the fraction of cache bytes eligible
for sharing, rather than the number of eligible layers alone. In a
controlled Gemma $K=4$ comparison, deduplication removes 58.5\% of
retained cache storage at approximately 1.8\% additional decode time.
Both the storage reduction and the update overhead are included in
the cost measurements below.

\subsection{Cost Evaluation}
\label{app:speed-protocol}

\paragraph{Analytic cost.}
\label{app:speed-analytic}
We summarize the cost trade-off between morphologies using a
static cost function. This provides a consistent training penalty
without introducing workload variation and measurement noise from
individual wall-clock observations:
\[
 C(m)=1+\frac{j-i}{N}k_{\mathrm{eff}}(\tau,\theta),\qquad
 k_{\mathrm{eff}}(\tau,\theta)=
 \begin{cases}
  0, & \tau=\basecomp{},\\
  r, & \tau=\mathrm{REPEAT},\\
  \kappa_K-1, & \tau=\mathrm{BRANCH}.
 \end{cases}
\]
For \rep{}, the intensity coefficient follows the number of additional passes. For \branch{}, we use $(\kappa_2,\kappa_4,\kappa_8)=(1.3,1.6,2.6)$ as effective penalty multipliers, informed by execution time and cache measurements of the optimized implementation. Parallel execution can accommodate additional branches with little change in decode throughput, while retaining distinct branch histories increases storage requirements. The multipliers incorporate both considerations into the training penalty. Together with the span-width factor, they define $C(m)$; subtracting one sets the penalty for \basecomp{} to zero. All coefficients remain fixed during selector training.

\paragraph{Measurement protocol.}
We measure all 840 held-out inputs on each backbone using  \texttt{bfloat16} weights, PyTorch 2.13.0+cu130, Transformers 5.15.1, and framework attention
implementations. Deduplication is enabled for every
strategy that selects \branch{}.

\begin{figure}[t]
\centering
\includegraphics[width=\textwidth]{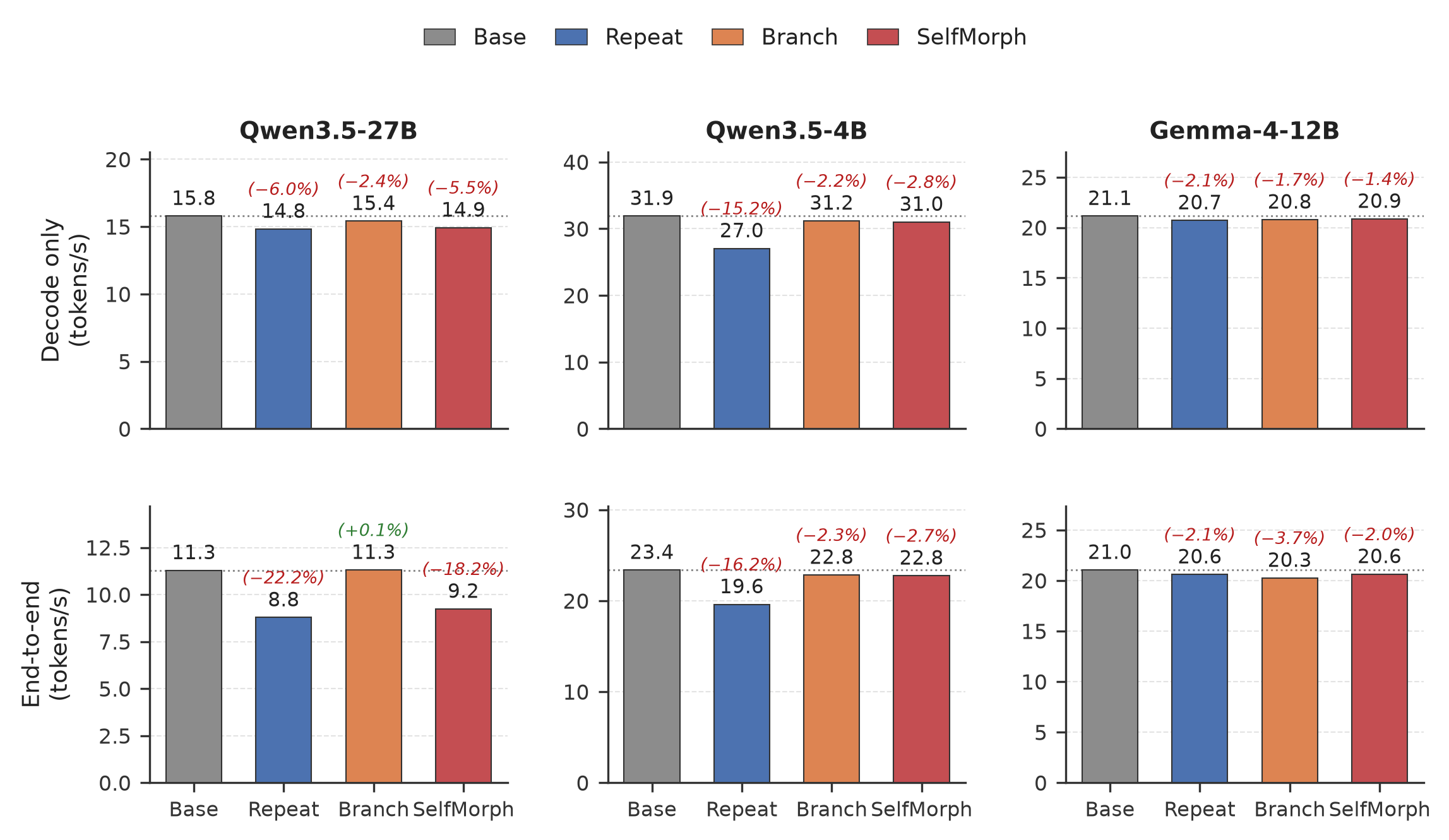}
\caption{Generation throughput on 840 inputs per backbone, with
branch-cache deduplication enabled. Top: decode-phase rate, excluding
prefill and its first output token. Bottom: generation-call throughput,
including prompt prefill. Bars show sample means; annotations compare
each mean with BASE. Strategies can produce different output lengths,
so throughput and per-query latency need not have the same ordering.}
\label{fig:time-cost}
\end{figure}

\begin{figure}[t]
\centering
\includegraphics[width=\textwidth]{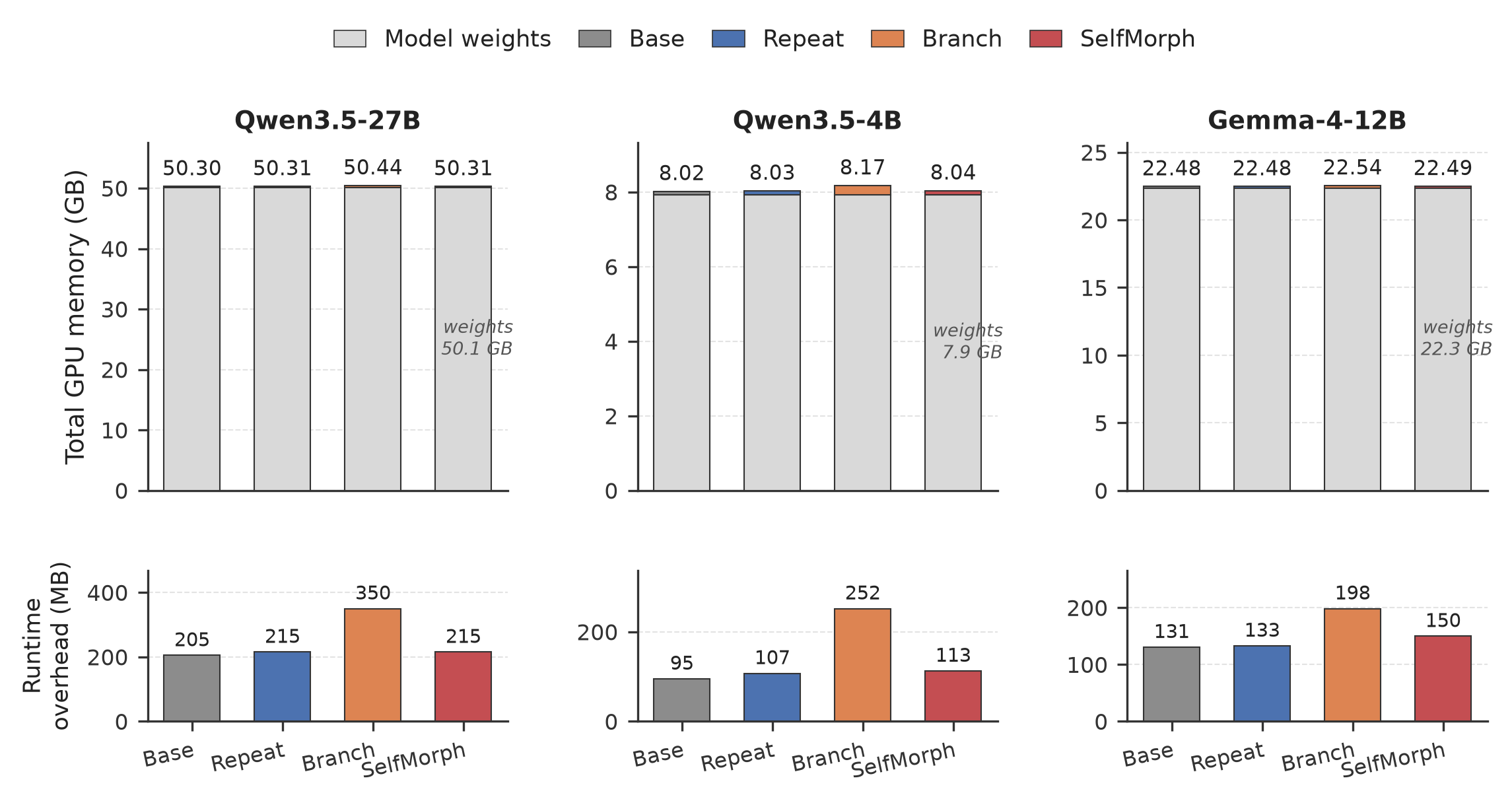}
\caption{Average resident GPU memory during generation on the same
840 inputs. Top: frozen weights and runtime overhead. Bottom: runtime
overhead on a separate scale. Eligible branch caches share physical
storage; recurrent states with incompatible in-place updates remain
replicated. Values are resident averages rather than peaks.}
\label{fig:memory-cost}
\end{figure}

Figures~\ref{fig:time-cost} and~\ref{fig:memory-cost} measure morphology
execution. For each strategy, the generation-call interval covers
prompt prefill and response decoding under its selected action.

Each input--strategy pair receives one warmup and two timed runs,
whose times are averaged. Strategy order rotates across inputs, and
allocator caches are retained between runs. Synchronization separates
prefill from decoding, with no per-token synchronization during
decoding. Let $t_{\mathrm{gen}}$ be the generation-call time and
$t_{\mathrm{prefill}}$ the first forward-pass time. For $n$ generated
tokens, the reported rates are
\[
 q_{\mathrm{decode}}
 =\frac{n-1}{t_{\mathrm{gen}}-t_{\mathrm{prefill}}},\qquad
 q_{\mathrm{gen}}=\frac{n}{t_{\mathrm{gen}}}.
\]
The first token is produced by prefill and excluded from the decode
numerator. Bars show sample-mean rates; annotations compare these
means with the corresponding BASE mean.

\paragraph{Memory accounting.}
Resident memory is sampled after each forward pass and averaged first
within a generation call, then across inputs. Subtracting frozen
weights gives runtime overhead, including caches and other runtime
allocations. Cache bytes count distinct backing allocations once,
including when several logical rows share storage. The figure reports
average resident memory; temporary peaks are a separate quantity.

\paragraph{Structural predictions and measured cost behavior.}
For \rep{}, the analytic model makes a directly testable prediction:
adding $r(j-i)$ layer evaluations increases the executed-depth ratio
to $C(m)$, implying a decode-rate ratio near $1/C(m)$ when layer costs
are comparable. Across the three fixed \rep{} configurations, the
measured decode-rate ratios differ from this prediction by less than
0.7 percentage points. Measured cache ratios also agree with $C(m)$
to within 0.8\% relative error. Thus, both execution time and storage
track the structural depth term across the three backbones.

For \branch{}, the measurements reveal distinct effects on decoding speed and storage. Across the three fixed branch configurations, decode throughput remains within 2.4\% of \basecomp{} (Figure~\ref{fig:time-cost}), consistent with amortized weight access during parallel execution. Additional branches retain distinct histories, increasing cache storage, while prefill cost depends on the expanded batch and prompt length. These complementary resource effects motivate considering both execution time and cache footprint when setting the proxy coefficients.

The cache-sharing model predicts the storage increase independently: for Gemma's span $(0,10)$ at $K=4$, uniform per-layer caches give $1+3(10/48)=1.625$ times single-replica storage, close to the measured ratio of approximately $1.661$. The byte-weighted formulation further accounts for unequal layer capacities and the recurrent states retained by hybrid backbones.

The Qwen strategies produce different output lengths under the same
limits. Their generation-call rates therefore reflect both execution
speed and the number of tokens over which prefill is amortized: a
shorter answer can finish sooner yet have lower tokens/s. Gemma has
matched output lengths in this workload, making the relationship
between its two timing panels more direct.

Figure~\ref{fig:memory-cost} isolates runtime overhead because frozen
weights dominate total memory. Gemma benefits most from branch-cache
sharing, consistent with its standard-attention caches. The hybrid
Qwen models retain replicated recurrent states, reducing the share of
storage that deduplication can remove. Together, the measurements show
how serial depth, parallel execution, and cache sharing contribute
differently to the cost of each morphology.

\section{Morphology Analysis}
\label{app:morphology-analysis}

This section presents two complementary views of how the morphology
types behave on frozen backbones: the geometric signature each type
leaves on the hidden state trajectory, and the task-level dissociation
between \rep{} and \branch{}.

\subsection{Trajectory Geometry}
\label{app:trajectory-viz}

\begin{figure}[h]
\centering
\includegraphics[width=0.6\linewidth]{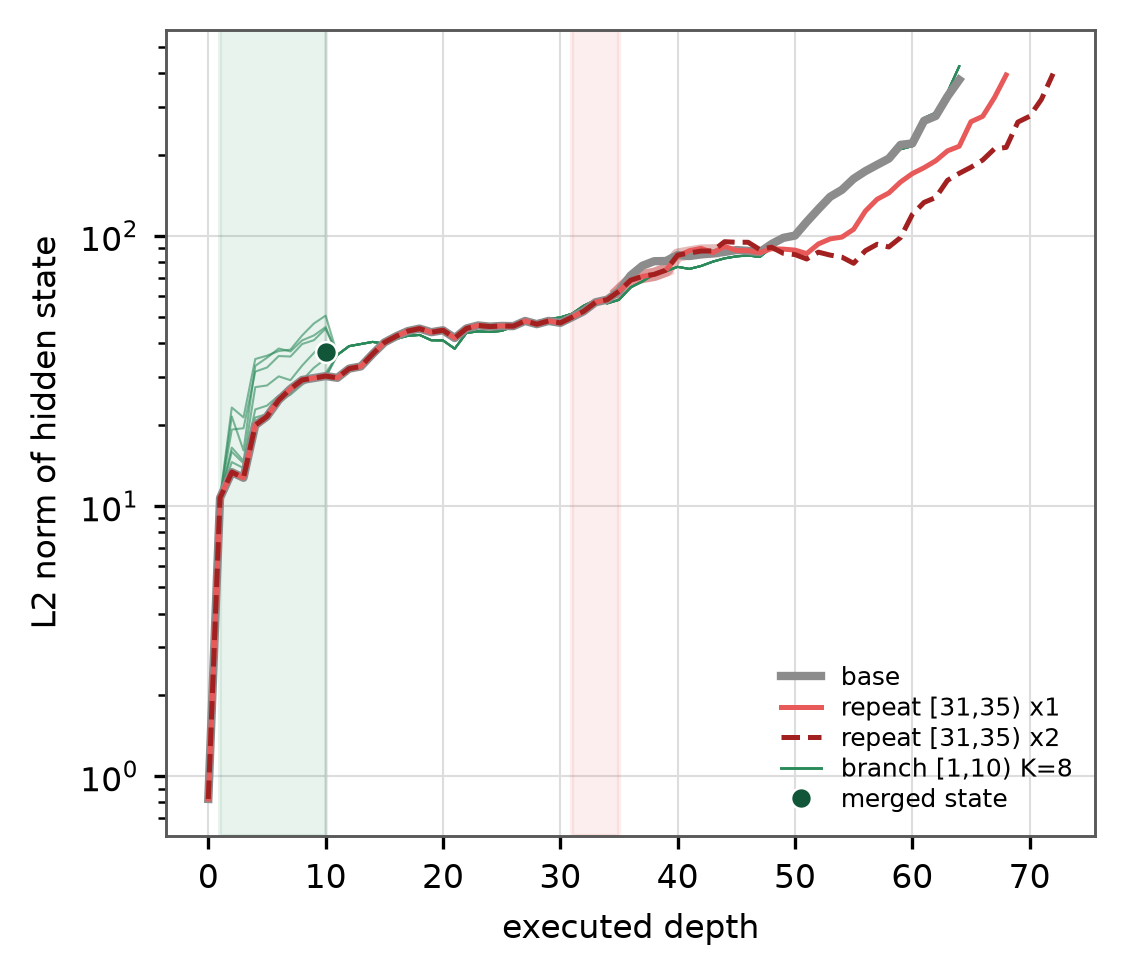}
\caption{Hidden state L2 norm (log scale) across executed depth for one
ARC-Challenge context (frozen Qwen3.5-27B). Gray: unmodified
\basecomp{} trajectory. Red solid/dashed: \rep{} with $r{=}1$ and
$r{=}2$ extra passes through span $[31,35)$ (shaded red).
Green: \branch{} with $K{=}8$ replicas through span $[1,10)$ (shaded
green). The eight thin lines fan out from the entry and collapse into one
merged state (dark dot) at the exit.}
\label{fig:trajectory-viz}
\end{figure}

Figure~\ref{fig:trajectory-viz} illustrates the effect of each
morphology type on the hidden state trajectory of a frozen Qwen3.5-27B
backbone, using an ARC-Challenge question.

The three morphology types produce qualitatively distinct trajectory
signatures. \basecomp{} (gray) follows the standard depth profile.
\rep{} (red) extends the trajectory past the depth the base forward
ever reaches: with $r{=}2$, the representation passes through the span
three times, with each traversal receiving the state produced by the preceding traversal. \branch{} (green) fans out at the span entrance into
$K$ diverging trajectories and collapses them into a single merged state
at the exit. The merged state resumes the standard depth profile from
that point, but at a different position in hidden space than the base
trajectory would have reached.

\subsection{Task-Level Dissociation}
\label{app:specialization}

The two non-\basecomp{} families help on different inputs. For every
training sample, we records the utility of all 25 actions. If a
single fixed action were always optimal, per-sample selection would be
unnecessary. The oracle distribution shows it is not.

\begin{figure}[h]
\centering
\includegraphics[width=\textwidth]{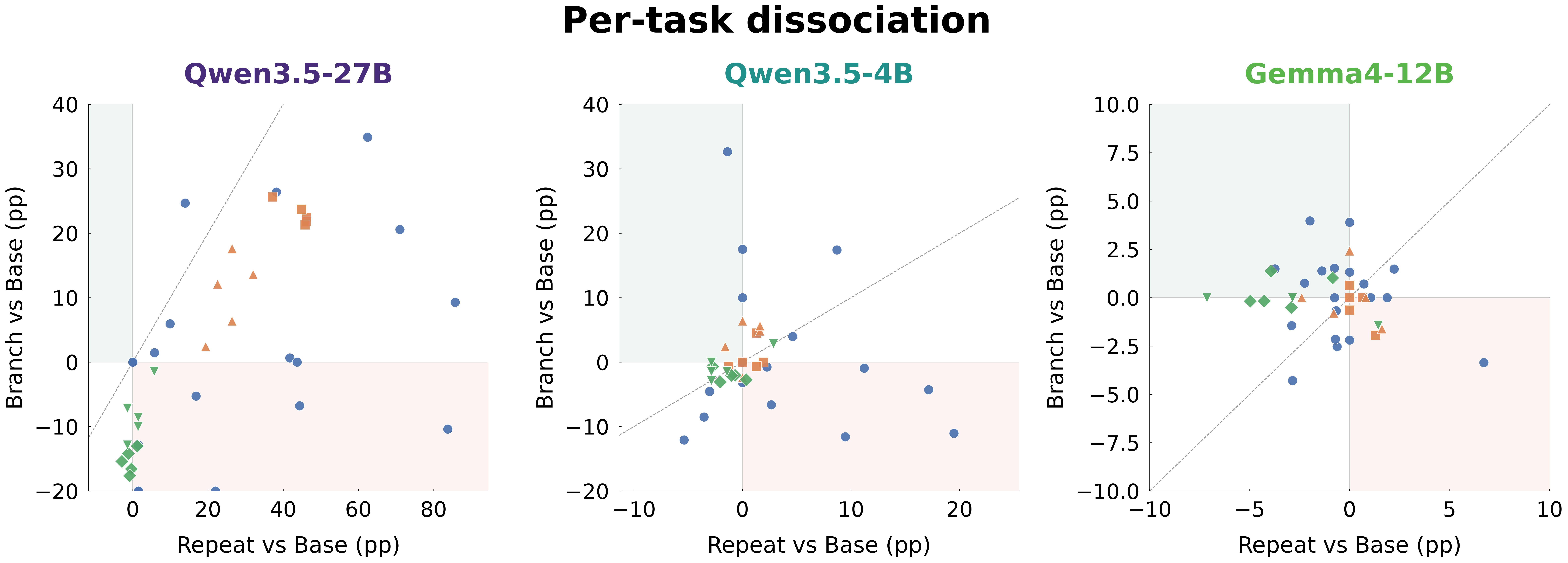}
\caption{Per-task dissociation between \rep{} and \branch{} on the training split. Each point is one BBH subtask
($\circ$) or one of five sample sub-blocks of ARC-Challenge
($\square$), OpenBookQA ($\triangle$), GSM8K ($\diamond$), or MATH-500
($\triangledown$), colored by task domain (blue = reasoning, orange =
science MC, green = mathematics). Axes show the net effect (help\%
$-$ hurt\%) relative to \basecomp{}. Points in the shaded off-diagonal
quadrants indicate that one morphology actively hurts while the other
helps on the same task group.}
\label{fig:dissociation}
\end{figure}

Figure~\ref{fig:dissociation} plots, for each task group, the net
effect of each backbone's single best \rep{} action ($x$-axis) against
that of its single best \branch{} action ($y$-axis). Points in the
shaded off-diagonal quadrants indicate tasks where one morphology
actively hurts while the other helps. On Qwen3.5-27B, some BBH subtasks improve under \rep{} but decline
under \branch{}, whereas many science sub-blocks improve under both.
Opposite-sign effects also occur on the smaller backbones, although
their magnitudes and task-group distributions differ.
These patterns show that the two fixed interventions have distinct
task-group effects.

\section{Additional Ablation Studies}
\label{app:additional-ablations}

We complement Section~\ref{sec:exp-ablations} with ablations of
selector architecture, capacity, distillation temperature, and cost
coefficient. Unless otherwise noted, ablations are conducted on
Qwen3.5-27B under the short-budget protocol, with all other
hyperparameters held fixed. Mean score denotes the average
per-sample task score.

\paragraph{Architecture components.}
Table~\ref{tab:ablation-arch} reports ablation results for removing two selector components. Self‑attention lets the query token aggregate cross‑layer trajectory information, weighting each depth by input‑dependent relevance so that highly informative layers shape the final decision vector. Pre-normalization applies the backbone's corresponding normalization
to the trajectory features. Removing either component lowers the
observed mean score, supporting their inclusion in the selector.

\begin{table}[t]
\centering
\caption{Architecture ablation. $\Delta$ is the absolute difference
in mean score from the default architecture.}
\label{tab:ablation-arch}
\small
\begin{tabular}{lcc}
\toprule
Variant & Mean Score & $\Delta$ \\
\midrule
Default        & \textbf{0.6083} &            \\
w/o PreNorm           & 0.6036          & $-0.0047$  \\
w/o Attention & 0.6012         & $-0.0071$  \\
\bottomrule
\end{tabular}
\end{table}

\paragraph{Capacity.}
Table~\ref{tab:ablation-cap} reports the effect of scaling the selector
across a 12$\times$ parameter range.
Among the evaluated configurations, the highest score occurs at
2.24M parameters. The deployed 1.05M setting performs close to this
configuration with fewer parameters. Both the smallest configuration
and those above 3M parameters obtain lower scores

\begin{table}[t]
\centering
\small
\begin{minipage}[t]{0.5\linewidth}
\centering
\captionof{table}{Capacity ablation on Qwen3.5-27B. Bold marks the deployed configuration.}
\label{tab:ablation-cap}
\begin{tabular}{rrrrc}
\toprule
$d$ & $d_z$ & $d_{\mathrm{head}}$ & Params & Mean Score \\
\midrule
96  & 48  & 24  & 528K  & 0.6036 \\
\textbf{192} & \textbf{64}  & \textbf{48}  & \textbf{1.05M} & \textbf{0.6083} \\
384 & 128 & 96  & 2.24M & 0.6095 \\
512 & 192 & 128 & 3.22M & 0.6060 \\
1024 & 256 & 256 & 6.44M & 0.6024 \\
\bottomrule
\end{tabular}
\end{minipage}\hfill
\begin{minipage}[t]{0.45\linewidth}
\centering
\captionof{table}{Temperature ablation on Qwen3.5-27B. Bold marks the deployed setting.}
\label{tab:ablation-temp}
\begin{tabular}{cc}
\toprule
$T$ & Mean Score \\
\midrule
0.01 & 0.6024 \\
0.10 & 0.6071 \\
\textbf{0.15} & \textbf{0.6083} \\
0.20 & 0.6036 \\
0.40 & 0.5964 \\
\bottomrule
\end{tabular}
\end{minipage}
\end{table}

\paragraph{Temperature.}
Table~\ref{tab:ablation-temp} varies the temperature $T$ of the
teacher distribution $q_T(m\mid x)$.
Among the tested values, $T{=}0.15$ achieves the highest score.
As described in Section~\ref{sec:method-objective}, lower temperatures
concentrate the teacher on higher-utility actions, while higher
temperatures increase its entropy and reduce probability differences
between actions. Exact utility ties retain equal probability at any
temperature. The results favor an intermediate temperature for this
backbone, balancing concentrated utility preferences with a broader
distribution over alternative actions.

\begin{figure}[t]
\centering
\includegraphics[width=\textwidth]{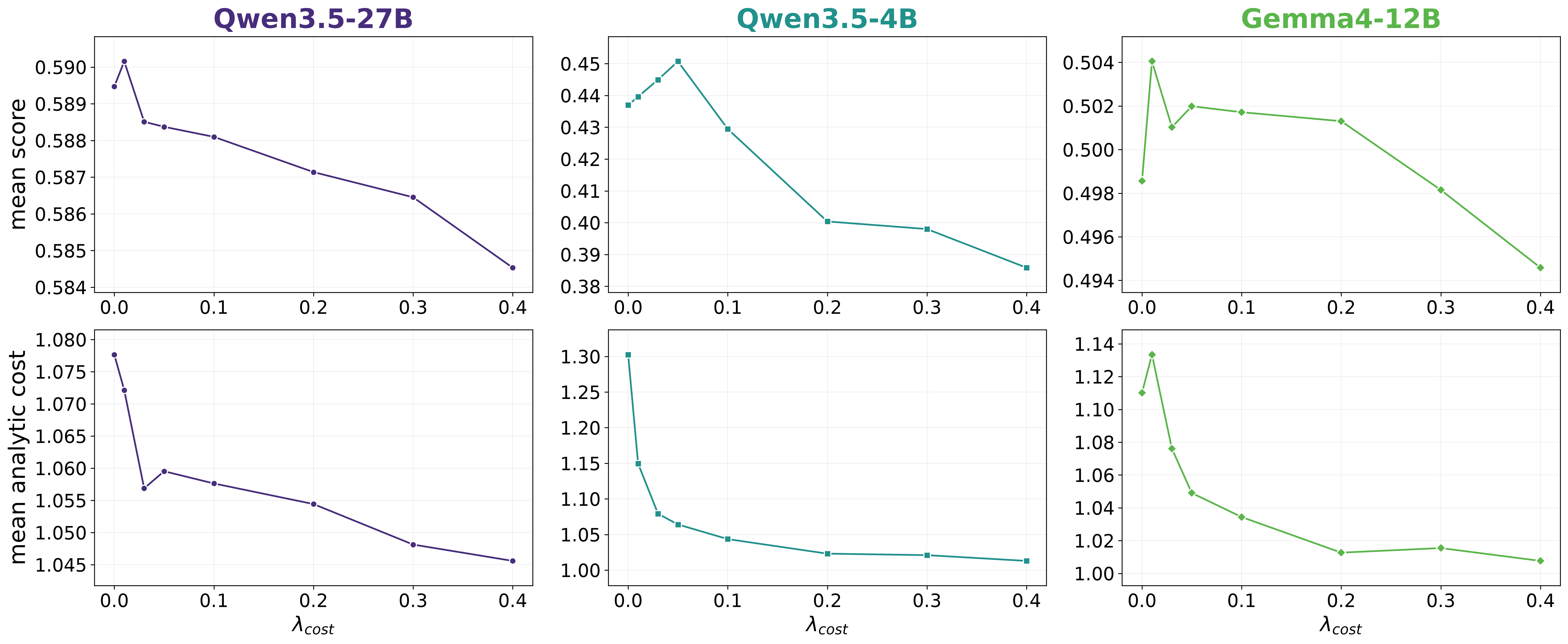}
\caption{Sensitivity to the cost coefficient $\lambda$ across three
backbones on the training split. Top: mean score. Bottom: mean analytic cost.
Each point represents a separately trained \method{}.}
\label{fig:lambda-pareto}
\end{figure}

\paragraph{Cost coefficient $\lambda$.}
Figure~\ref{fig:lambda-pareto} shows the effect of the cost
coefficient $\lambda$ in the utility
$U(x,m)=\mathrm{score}(x,m)-\lambda(C(m)-1)$.
Mean analytic cost generally decreases as $\lambda$ increases.
Increasing $\lambda$ strengthens the teacher's preference for
lower-cost actions. A small positive coefficient improves mean score
over $\lambda=0$ on all three backbones, while larger coefficients
generally reduce score. For actions with equal task scores, the cost
term favors the less expensive alternative. The observed gains suggest
that this preference can also benefit selection in the evaluated setting.

%% file: Sections/background.tex
\section{Related Work}
\label{sec:related_work}

\paragraph{Recurrent computation.}
Recurrent-depth (looped) Transformers organize computation through repeated application of shared transformations. Universal Transformers introduce recurrence across depth \citep{dehghani2019universal}; subsequent work trains models with variable unrolled lengths for inference-time scaling \citep{geiping2026scaling} and extends this paradigm to large-scale pretraining \citep{zhu2025scaling}. These models are trained to support recurrent execution. Another line reconfigures pretrained layers while keeping the backbone frozen. PoLar learns input-dependent programs that skip or repeat layer modules \citep{li2026skip}, while training-free looped Transformers repeat middle-layer blocks and stabilize their updates without parameter learning \citep{chen2026training}. \method{} shares the goal of adapting execution over pretrained weights. Its focus is the joint selection of serial reuse and local branching, including their spans and intensities, using the backbone's prompt trajectory to predict which intervention is useful.

\paragraph{Parallel computation.}
Parallel methods expand computation through multiple latent streams. \citet{chen2026parallel} learn input transformations and an aggregation mechanism for parallel model executions. SoftCoT++ constructs multiple latent candidates at the token or embedding interface \citep{xu2025softcot++}, while Multiplex Thinking branches over sampled tokens and merges their embeddings during continuous reasoning \citep{tang2026multiplex}. Other approaches sample continuous reasoning trajectories from a latent-reasoning backbone and train a reward model to select among them \citep{you2026parallel}. These methods differ in where streams diverge, how they are combined, and which components are learned. \method{} implements branching within a selected layer span of a frozen backbone, using Gaussian perturbations and arithmetic averaging without learning the perturbation or merge operation. Its central question is when this local branching is preferable to serial layer reuse or the original forward path.